\documentclass[a4paper,fleqn]{cas-dc}

\usepackage[authoryear,longnamesfirst]{natbib}
\usepackage{times}
\usepackage{soul}
\usepackage{url}
\usepackage[small]{caption}
\usepackage{graphicx}
\usepackage{amsmath}
\usepackage{amsthm}
\usepackage{booktabs}
\usepackage{algorithm}
\usepackage{algorithmic}
\usepackage[switch]{lineno}

\usepackage{amssymb}
\usepackage{paralist}

\usepackage{adjustbox}
\usepackage{multirow}
\usepackage{makecell}
\usepackage{booktabs}
\usepackage{float}
\usepackage[table]{xcolor}
\def\tsc#1{\csdef{#1}{\textsc{\lowercase{#1}}\xspace}}
\tsc{WGM}
\tsc{QE}

\newtheorem{theorem}{Theorem}
\newtheorem{lemma}[theorem]{Lemma}
\newdefinition{rmk}{Remark}
\newproof{pf}{Proof}
\newproof{pot}{Proof of Theorem \ref{thm}}
\newtheorem{proposition}{Proposition}
\newcommand{\pinkcell}[1]{\cellcolor{pink!20}{#1}}

\begin{document}
\let\WriteBookmarks\relax
\def\floatpagepagefraction{1}
\def\textpagefraction{.001}

\shorttitle{}    

\shortauthors{}  

\title [mode = title]{Outlier Smoothing with Closed-Form Rotations for W4A4 Large Language Model Quantization}  



%
\author[1]{Jinying Xiao}
\credit{}



\ead[url]{}
\author[1]{Bin Ji}
\credit{}
\author[1]{Shasha Li}
\credit{}
\author[1]{Xiadong Liu}
\credit{}
\author[1]{Jun Ma}
\credit{}
\author[2]{Chao Wang}
\credit{}
\author[1]{Wei Li}
\credit{}
\author[1]{Ye Zhong}
\credit{}
\author[1]{Xuan Xie}
\credit{}
\author[3]{Nyima Tashi}
\credit{}
\author[1]{Jie Yu}
\credit{}
\cormark[1]


\affiliation[1]{organization={National University of Defense Technology},
            addressline={Deya Road}, 
            city={Changsha},
            postcode={410073}, 
            state={Hunan},
            country={China}}
\affiliation[2]{organization={Qinghai Normal University},
            addressline={Wusi Street}, 
            city={Xining},
            postcode={810016}, 
            state={Qinghai},
            country={China}}
\affiliation[3]{organization={Tibet University},
            addressline={Zangda East Road}, 
            city={Lasa},
            postcode={850000}, 
            state={Xizang},
            country={China}}









\begin{abstract}
Post-training quantization (PTQ) is a practical approach for reducing the memory footprint and inference cost of large language models (LLMs), with W4A4 weight-activation quantization offering substantial potential for efficient deployment. However, extremely low-bit activation quantization remains challenging due to severe outlier effects. In particular, massive outliers and normal outliers distort activation distributions, dominate the quantization range, and reduce effective quantization-space utilization. Existing rotation-based methods mitigate this issue by learning orthogonal transformations, but their reliance on gradient-based optimization introduces substantial calibration overhead and is poorly aligned with the non-differentiable nature of quantization. In this paper, we propose SingleQuant, an optimization-free rotation framework for W4A4 PTQ. We first analyze the instability of STE-based rotation learning and show that quantization-induced discontinuities can lead to non-smooth Riemannian gradients and persistent Cayley-update oscillations. Motivated by this analysis, SingleQuant constructs rotations directly through closed-form Givens transformations. Specifically, Alignment Rotation Transformation smooths sparse massive outliers with locally optimal two-dimensional rotations, while Uniformity Rotation Transformation reshapes normal outliers toward a more quantization-friendly distribution. A Kronecker-structured design further reduces the cost of applying large orthogonal transformations. Extensive experiments across multiple LLM architectures and downstream benchmarks demonstrate that SingleQuant achieves competitive W4A4 accuracy, significantly reduces quantization time, and preserves inference efficiency, providing a deterministic and practical alternative to optimization-based rotation quantization.
\end{abstract}





\maketitle

\begin{figure*}[t] 
	\centering
	\includegraphics[width=0.85\textwidth]{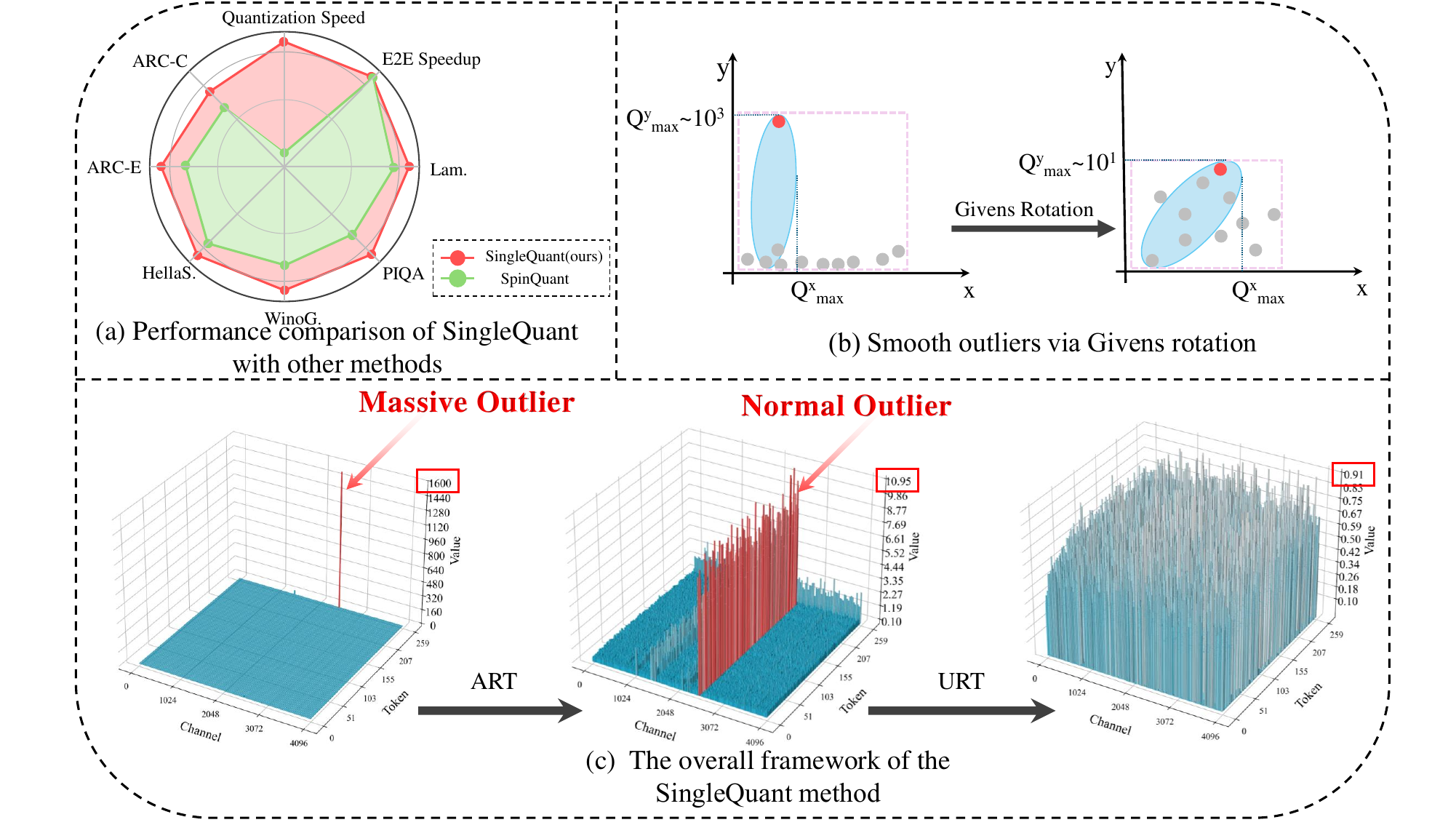} 
	\caption{The sub-figure (a) compares SingleQuant and SpinQuant in quantization speed (LLMs quantization per hour), end-to-end speedup, and performance under various QA tasks. The sub-figure (b) illustrates deterministic outlier smoothing via Givens rotation on 2D data containing MO. Grey circles represent data points, with red ones indicating MO. Blue ellipses depict quantization space (size determined by bit-width), where greater coverage of data points within this fixed space indicates higher quantization space utilization. The sub-figure (c) presents SingleQuant's framework comprising two components: ART smooths outlier magnitudes targeting prominent/scattered outliers, while URT performs secondary smoothing through distribution optimization. The diagram demonstrates ART/URT operations against distinct outlier types.}
	\label{overall}
\end{figure*}

\section{Introduction}
LLMs have demonstrated strong capabilities across a wide range of scientific tasks \citep{aggarwal2026large,huang2025large}. However, performing inference with LLMs requires substantial computational and storage resources. Reducing the resource requirements of LLM inference has therefore received considerable research attention, with post-training quantization (PTQ) \citep{ashkboos2023towards,lin2024duquant,zhao2024atom,maaffinequant,liu2025fbquant} emerging as a promising solution.

Among various PTQ settings, W4A4 weight-activation quantization has emerged as a particularly important scenario for efficient LLM deployment, as it simultaneously reduces weight storage and activation memory bandwidth, enabling more practical low-bit inference on modern hardware \citep{xiao2023smoothquant,ashkboos2024quarot,liu2024spinquant}. However, despite its strong efficiency potential, W4A4 quantization remains substantially more challenging than weight-only quantization because low-bit activations are highly sensitive to input-dependent distributional irregularities. In particular, activation outliers, including massive outliers (MO) and normal outliers (NO), remain a major unresolved challenge in PTQ settings \citep{lin2024duquant,jinmassive,ramachandran2025microscopiq}. These outliers dominate the quantization range and reduce the effective bit allocation for most values, leading to poor utilization of the quantization space \citep{huostquant}. To mitigate this issue, prior work has exploited rotational invariance to redistribute outlier energy and improve model quantizability, typically by optimizing rotation matrices with gradient-based methods \citep{sunflatquant,liu2024spinquant,huostquant}.

\begin{figure}
\centerline{\includegraphics[width=0.5\textwidth]{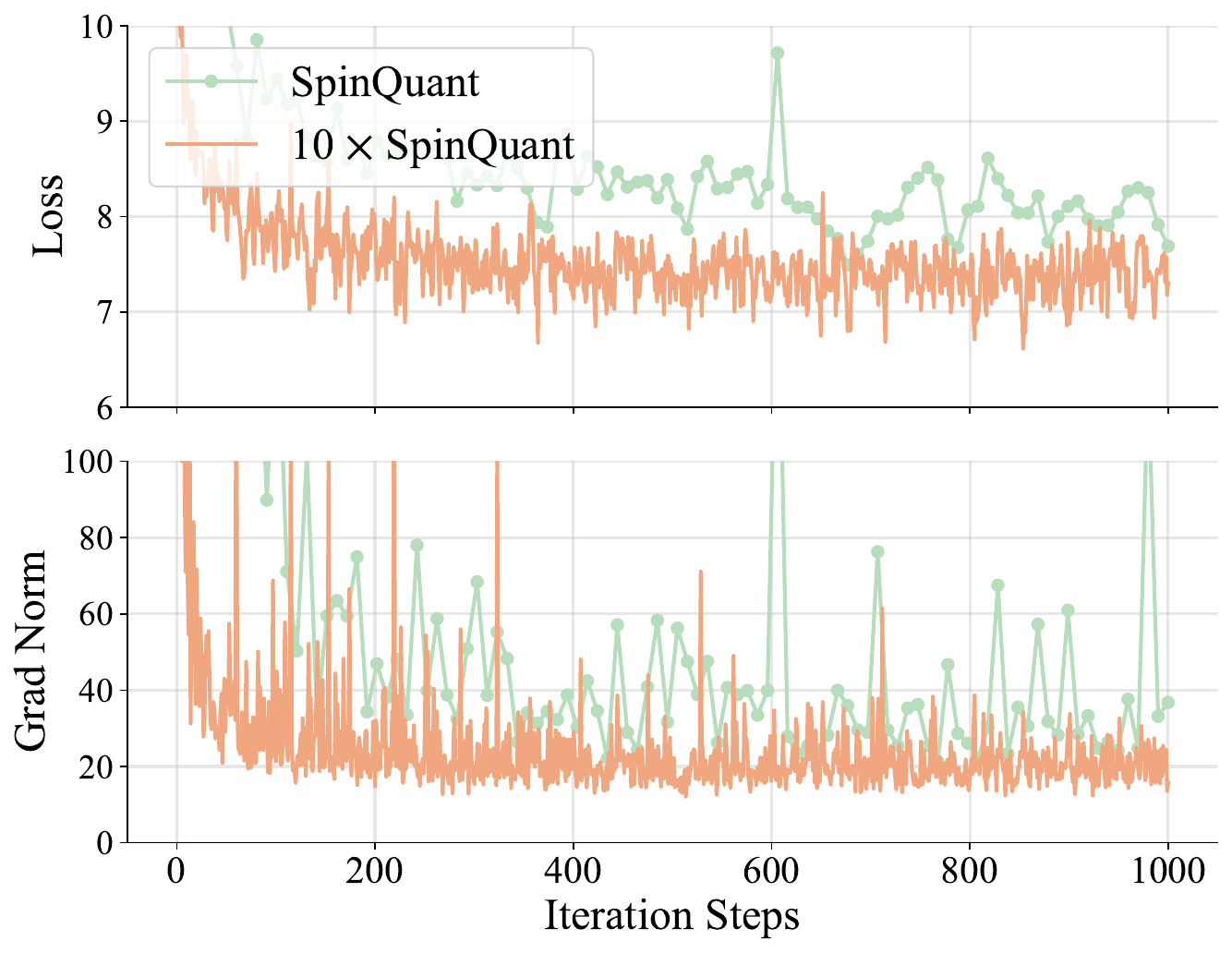}}
	\caption{SpinQuant applies W4A4 quantization to LLaMA-2-7B with linearly decaying LR. The orange curve uses 10$\times$ SpinQuant’s claimed iterations; adjacent green points are spaced 10 iterations apart. The figure shows optimization loss and gradient norm. More detailed results are provided in Appendix~Fig.~\ref{app:spinquant_loss}.}
    \label{spinquant_loss}
\end{figure}

However, we observe that gradient-based optimization is poorly aligned with discrete quantization operations, leading to pathological convergence behavior (Fig.~\ref{spinquant_loss}). In addition, gradient-based adjustment of rotation matrices introduces substantial time overhead (Fig.~\ref{overall}a) and remains dependent on GPTQ support. Our theoretical analysis shows that Cayley SGD combined with the Straight-Through Estimator (STE) \citep{liefficient} exhibits non-smooth and oscillatory convergence behavior. Specifically, the STE introduces persistent noise relative to the true gradients \citep{bengio2013estimating}, which causes optimization oscillations and hinders convergence. Moreover, the STE induces non-smooth changes across quantization boundaries, violating the Lipschitz smoothness condition required for the convergence of Cayley SGD and thereby preventing gradient stabilization. As a result, SpinQuant is affected by pathological optimization dynamics and consistently produces suboptimal results, as verified empirically.

These limitations hinder the broader adoption of quantization techniques. To address this issue, we propose SingleQuant (see Fig.~\ref{overall}c), a mathematically constructed framework that is decoupled from quantization operations and does not require optimization, thereby making quantization a deterministic computation. The key idea is to use the geometric properties of Givens rotations \citep{press2007numerical} to directly address both types of outliers. Alignment Rotation Transformation (ART) targets sparse but extreme-valued MO by smoothing extreme values in a single operation using closed-form optimal rotation angles. As shown in Fig.~\ref{overall}b, ART aligns outliers within the quantization domain and substantially improves quantization-space utilization. Uniformity Rotation Transformation (URT) addresses widely distributed NO with moderate magnitudes by reshaping the activation distribution to remove quantization bottlenecks. In particular, URT leverages the feasibility of Givens mapping \citep{ma2024parameter} to construct a uniform distribution mapping through linearly scaling Givens rotations, thereby reducing quantization interference induced by NO.

Both ART and URT are built on rigorously formalized Givens rotations. The main design novelty is that the rotation dimensions and angles are determined by the magnitude and distribution of outliers. Therefore, all critical parameters in the quantization process are obtained from mathematically derived closed-form solutions, rather than from gradient feedback on the Stiefel manifold, which may lead to pathological convergence in conventional methods. As shown in Fig.~\ref{overall}a, SingleQuant achieves a favorable trade-off among task performance, inference latency, and quantization speed compared with the evaluated baselines. Extensive experiments across diverse tasks and LLMs further demonstrate that SingleQuant consistently improves quantization efficiency while maintaining strong task performance under the considered W4A4 settings.

The contributions of this work are summarized below:
\begin{itemize}
	\item[(1)] 
    We analyze the optimization difficulty of STE-based rotation learning for low-bit weight-activation quantization. We show that the STE-induced Riemannian gradient estimator can become non-smooth near quantization boundaries, and that persistent STE residuals may lead to non-vanishing Cayley updates. This analysis explains the oscillatory behavior observed in optimization-based rotation methods and motivates an optimization-free rotation construction.
	\item[(2)] We propose SingleQuant, a closed-form rotation framework for W4A4 post-training quantization of LLMs. SingleQuant explicitly handles two types of outliers through complementary Givens-rotation components: ART smooths sparse massive outliers with locally optimal two-dimensional rotations, while URT reshapes normal outliers toward a norm-preserving uniform target. In addition, Kronecker decomposition is introduced to reduce the cost of applying large orthogonal transformations.
	\item[(3)] Experimental results show that SingleQuant provides competitive performance under the evaluated W4A4 quantization settings. For example, on LLaMA-3-70B, SingleQuant attains 76.30\% average zero-shot accuracy, exceeding the strongest evaluated baseline by 3.33 percentage points under the same benchmark suite. In terms of quantization time, SingleQuant quantizes LLaMA-2-13B in 37 seconds, corresponding to an approximately $1400\times$ reduction compared with the 14-hour runtime reported for the optimization-based baseline.

\end{itemize}

\section{Related Work}
\textbf{LLMs Quantization.}
Quantization is a key technique for reducing the memory footprint, bandwidth cost, and inference latency of LLMs by representing weights and activations with low-bit values. Among existing approaches, PTQ is particularly attractive because it compresses pretrained models without expensive retraining. However, low-bit quantization for LLMs remains challenging, especially under weight-activation quantization, where activation distributions are input-dependent and highly sensitive to outliers.

Prior studies have shown that LLMs exhibit salient activation outliers \citep{shen2020q,bai2021binarybert}, as well as massive outliers (MO) associated with pivot tokens and attention-related representations \citep{wei2022outlier,barberoround,liu2024intactkv,lin2024duquant,sun2024massive,jinmassive}. These extreme values dominate the quantization range, causing most normal values to occupy only a small portion of available quantization levels and thereby degrading model accuracy. To mitigate this issue, pre-quantization transformations are widely used to reshape weights and activations while preserving the full-precision computation. Scaling-based methods exploit channel-wise invariance to balance activation and weight magnitudes \citep{xiao2023smoothquant,maaffinequant}, whereas rotation-based methods use orthogonal transformations to redistribute outlier energy across dimensions \citep{ashkboos2024quarot}.

\textbf{Learnable Transformation.}
Beyond fixed pre-quantization transformations, recent studies have explored learnable transformations and clipping thresholds to further adapt LLMs to low-bit quantization. These methods typically introduce trainable scaling factors, rotation matrices, or clipping parameters that are optimized on calibration data to reshape activation and weight distributions before quantization. By directly minimizing quantization-induced reconstruction error, learnable transformations can provide stronger adaptability than hand-crafted or random transformations.

Among them, SpinQuant \citep{liu2024spinquant} learns orthogonal rotations for LLM quantization and maintains the orthogonality constraint through Cayley SGD, enabling better outlier redistribution while preserving the full-precision computation. DuQuant \citep{lin2024duquant} accelerates rotation search with a greedy learning strategy and dual transformations, reducing calibration cost but still showing limited performance under challenging low-bit settings. FlatQuant \citep{sunflatquant} applies online matrix transformations to flatten weight and activation distributions, improving quantization robustness at the expense of additional inference overhead and extra transformation parameters. OSTQuant \citep{huostquant} further combines learnable orthogonal and scaling transformations to better fit quantized distributions, achieving strong performance across multiple LLM benchmarks. Despite these advances, learnable transformation methods usually require iterative calibration or gradient-based optimization, which increases quantization time and may be sensitive to optimization instability caused by non-differentiable quantization operations. 

\section{Discussions}
\subsection{Information-Preserving Outlier Smoothing}
In low-bit weight-activation quantization, activation outliers are a key source of quantization-induced information loss~\citep{jinmassive,lin2024duquant}. As shown in Fig.~\ref{overall}c, activation distributions in LLMs are input-dependent and often exhibit clear long-tailed patterns, where a small number of channels or tokens contain values much larger than ordinary activations. These MO greatly enlarge the activation dynamic range, while NO with moderate magnitudes but broader distributions continuously affect quantization errors across multiple channels. Since quantization discretizes numerical values in a coordinate space, a scale dominated by MO forces the finite low-bit range to cover a few extreme values, leaving most normal activations with only a small number of effective quantization levels. As a result, many low-magnitude but informative activations are mapped to identical or nearby quantized values, and originally distinguishable full-precision states collapse into low-bit representations that are difficult to separate. For deep Transformers, this information distortion propagates through attention, MLP, and residual branches, ultimately leading to poor performance under extremely low-bit settings such as W4A4.

Recent studies ~\citep{ashkboos2024quarot,sunflatquant,liu2024spinquant} use orthogonal rotation matrices to smooth activations by reorganizing the distribution of information across coordinate dimensions while preserving the original information structure. Specifically, for an orthogonal matrix $\mathbf{R}$, an activation vector $\mathbf{x}$ satisfies $\|\mathbf{x}\mathbf{R}\|_2=\|\mathbf{x}\|_2$, and any two activation vectors $\mathbf{x}_1$ and $\mathbf{x}_2$ also satisfy $(\mathbf{x}_1\mathbf{R})(\mathbf{x}_2\mathbf{R})^\top=\mathbf{x}_1\mathbf{x}_2^\top$. Therefore, rotation does not change the energy, similarity, or geometric relations of full-precision representations. In a linear layer, computational equivalence can also be preserved by rotating activations and weights simultaneously:
\begin{equation}
\mathbf{X}\mathbf{W} = (\mathbf{X}\mathbf{R})(\mathbf{R}^{\top}\mathbf{W}) .
\end{equation}
A rotation matrix is thus an information-preserving coordinate transformation. Orthogonal rotation redistributes outlier energy across multiple dimensions, reduces the maximum coordinate magnitude, and makes the distribution more balanced. This allows finite low-bit quantization levels to encode the main differences in the original activations more effectively, thereby improving the information fidelity of quantized representations.
\subsection{Instability of STE-Based Rotation Learning}
While prior work~\citep{liu2024spinquant} employs Cayley SGD to optimize
rotation matrices on the Stiefel manifold, we observe that such gradient-based
optimization can become unstable when combined with quantization operations.
As shown in Fig.~\ref{spinquant_loss}, SpinQuant exhibits persistent oscillations in both the
objective value and the STE gradient norm, even when the number of optimization
iterations is increased by \(10\times\). In this section, we explain this
phenomenon from two perspectives: first, the STE-induced Riemannian gradient
estimator can fail to satisfy Lipschitz continuity near quantization boundaries;
second, the STE tangent estimator can have a non-vanishing update floor under
finite-step Cayley updates.

Let $\mathbf X\in\mathbb R^{N\times n}$ denote calibration activations, where $N$ is the number of calibration tokens
or samples and $n$ is the hidden dimension. Let $\mathbf W\in\mathbb R^{n\times C}$
denote a weight matrix with output dimension $C$. We optimize an orthogonal
rotation matrix
\begin{equation}
\mathbf R\in\mathcal M,\quad\mathcal M:=O(n)=\{\mathbf R\in\mathbb R^{n\times n}:\mathbf R^\top \mathbf R=\mathbf I_n\}.
\end{equation}
The tangent space of $\mathcal{M}$ at $\mathbf{R}$ is
\begin{equation}
T_{\mathbf R}\mathcal M=\{\boldsymbol\Omega \mathbf R:\boldsymbol\Omega^\top=-\boldsymbol\Omega\}.
\end{equation}
For any ambient matrix $\mathbf A\in\mathbb R^{n\times n}$, the orthogonal
projection onto $T_{\mathbf R}\mathcal M$ is denoted by
\begin{equation}
\Pi_{\mathbf R}(\mathbf A)=\mathbf A-\mathbf R\,\mathrm{sym}(\mathbf R^\top\mathbf A),\quad\mathrm{sym}(\mathbf B):=\frac{1}{2}(\mathbf B+\mathbf B^\top).
\end{equation}

Let $\mathbf Z:\mathcal M\rightarrow \mathbb R^m$
be a smooth pre-quantization mapping depending on $\mathbf R$. For example,
$\mathbf Z(\mathbf{R})$ may be the vectorized form of the rotated activations
$\mathbf X\mathbf{R}$, the rotated weights $\mathbf R^\top\mathbf{W}$, or
their concatenation. Quantization is applied to these rotated tensors rather
than to the invariant full-precision product
\begin{equation}
\mathbf X\mathbf R\mathbf R^\top\mathbf W=\mathbf X\mathbf W.
\end{equation}

Let $Q_\Delta:\mathbb R\rightarrow \Delta\mathbb Z$ be the scalar uniform
rounding quantizer with step size $\Delta>0$:
\begin{equation}
Q_\Delta(z)
=
\Delta\left\lfloor \frac{z}{\Delta}+\frac{1}{2}\right\rfloor .
\end{equation}
The same notation $Q_\Delta(\cdot)$ is used for its component-wise extension
to vectors. Its quantization boundaries are
\begin{equation}
\mathcal B_\Delta=\left\{\left(k+\frac{1}{2}\right)\Delta:k\in\mathbb Z\right\}.
\end{equation}
A representative quantization-aware surrogate objective is
\begin{equation}
\mathcal L_\Delta(\mathbf R)=\frac{1}{2}\left\|Q_\Delta(\mathbf Z(\mathbf R))-\mathbf Y\right\|_2^2,
\end{equation}
where $\mathbf Y\in\mathbb R^m$ denotes the corresponding full-precision
target.

Since $Q_\Delta(\cdot)$ is piecewise constant and non-differentiable, the
straight-through estimator (STE) replaces the derivative of the quantizer with
the identity map during backpropagation. Under this convention, the Euclidean
STE gradient estimator is
\begin{equation}
\mathbf G_{\mathrm E}^{\mathrm{STE}}(\mathbf R)=D\mathbf Z(\mathbf R)^*\left(Q_\Delta(\mathbf Z(\mathbf R))-\mathbf Y\right),
\end{equation}
where $D\mathbf Z(\mathbf R)^*$ denotes the adjoint of the differential
\(D\mathbf Z(\mathbf R)\). The corresponding Riemannian STE gradient estimator is
\begin{equation}
\mathbf G_{\mathcal M}^{\mathrm{STE}}(\mathbf R)=\Pi_{\mathbf R}\left(\mathbf G_{\mathrm E}^{\mathrm{STE}}(\mathbf R)\right).
\end{equation}

\begin{proposition}\label{pro1}
Suppose there exist
\(\mathbf R_0\in\mathcal M\), a coordinate index \(\rho\in\{1,\ldots,m\}\), and
a tangent direction \(\boldsymbol\xi\in T_{\mathbf R_0}\mathcal M\) such that
\begin{equation}
Z_\rho(\mathbf R_0)=\tau\in\mathcal B_\Delta
\end{equation}
and
\begin{equation}
D Z_\rho(\mathbf R_0)[\boldsymbol\xi]\neq 0.
\end{equation}
Assume further that the corresponding projected gradient jump is nonzero:
\begin{equation}
\mathbf v:=\Pi_{\mathbf R_0}\left[D\mathbf Z(\mathbf R_0)^*\mathbf e_\rho\right]\neq \mathbf 0,
\end{equation}
where \(\mathbf e_\rho\in\mathbb R^m\) is the \(\rho\)-th standard basis vector.
Then the mapping
\begin{equation}
\mathbf R\mapsto \mathbf G_{\mathcal M}^{\mathrm{STE}}(\mathbf R)
\end{equation}
is not Lipschitz continuous in any neighborhood of \(\mathbf R_0\). Consequently,
there is no global constant \(L>0\) such that
\begin{equation}
\left\|\mathbf G_{\mathcal M}^{\mathrm{STE}}(\mathbf R_1)-\mathbf G_{\mathcal M}^{\mathrm{STE}}(\mathbf R_2)\right\|_F\leq L\|\mathbf R_1-\mathbf R_2\|_F
\end{equation}
for all $\mathbf R_1,\mathbf R_2\in\mathcal M$.
\end{proposition}
The proof is presented in Appendix~\ref{app:a.1}.

Proposition~\ref{pro1} shows that the smoothness condition required by standard
Cayley-SGD convergence analyses may fail in quantization-aware rotation
learning. We next show that, under finite-step Cayley updates, a persistent STE
error component produces a non-vanishing update floor.

\begin{proposition}
\label{prop:persistent-ste-cayley}
Let $\mathcal{M}=O(n)$ and let $\{R_t\}_{t\ge 0}\subset \mathcal{M}$ be generated by the Cayley update
\begin{equation}
R_{t+1}
=
\left(I-\frac{\alpha_t}{2}\Omega_t\right)^{-1}
\left(I+\frac{\alpha_t}{2}\Omega_t\right)R_t ,
\label{eq:cayley-update-revised}
\end{equation}
where $\alpha_t>0$, $\Omega_t^\top=-\Omega_t$, and
\begin{equation}
\Omega_t=-\widehat{G}_tR_t^\top .
\label{eq:omega-def-revised}
\end{equation}
Here $\widehat{G}_t\in T_{R_t}\mathcal{M}$ denotes the tangent gradient estimator produced by the straight-through estimator (STE).

Assume that $\widehat{G}_t$ admits the decomposition
\begin{equation}
\widehat{G}_t = H_t+\varepsilon_t ,
\label{eq:ste-decomposition}
\end{equation}
where $H_t\in T_{R_t}\mathcal{M}$ is a reference tangent direction, e.g., the Riemannian gradient of a smooth surrogate objective, and $\varepsilon_t\in T_{R_t}\mathcal{M}$ is the residual component induced by the STE approximation and quantization mismatch. Let $\mathcal{F}_t$ denote the filtration generated by the optimization trajectory up to iteration $t$.

Suppose that there exist constants $\sigma>0$, $\kappa\in[0,1)$, and $G>0$ such that
\begin{equation}
\mathbb{E}\!\left[
\|\varepsilon_t\|_F^2 \mid \mathcal{F}_t
\right]
\ge \sigma^2 ,
\label{eq:residual-floor}
\end{equation}
\begin{equation}
\mathbb{E}\!\left[
\langle H_t,\varepsilon_t\rangle \mid \mathcal{F}_t
\right]
\ge
-\frac{\kappa}{2}
\mathbb{E}\!\left[
\|\varepsilon_t\|_F^2 \mid \mathcal{F}_t
\right],
\label{eq:residual-noncancellation}
\end{equation}
and
\begin{equation}
\|\Omega_t\|_2\le G
\quad \text{almost surely}.
\label{eq:omega-bound}
\end{equation}
Then
\begin{equation}
\mathbb{E}\!\left[
\|\widehat{G}_t\|_F^2 \mid \mathcal{F}_t
\right]
\ge
(1-\kappa)\sigma^2 .
\label{eq:estimator-floor}
\end{equation}

Moreover, if $\alpha_t\le \overline{\alpha}$ for all $t$, then
\begin{equation}
\mathbb{E}\!\left[
\|R_{t+1}-R_t\|_F^2 \mid \mathcal{F}_t
\right]
\ge
\frac{\alpha_t^2}{1+(\overline{\alpha}G/2)^2}
(1-\kappa)\sigma^2 .
\label{eq:cayley-displacement-lower}
\end{equation}
Consequently, if the step sizes are bounded away from zero, i.e.,
$\overline{\alpha}\ge\alpha_t\ge \underline{\alpha}>0$, then
\begin{equation}
\liminf_{t\to\infty}
\mathbb{E}\!\left[
\|R_{t+1}-R_t\|_F^2
\right]
\ge
\frac{\underline{\alpha}^2}{1+(\overline{\alpha}G/2)^2}
(1-\kappa)\sigma^2
>0 .
\label{eq:nonvanishing-displacement}
\end{equation}
If instead $\alpha_t\to0$, then the non-vanishing conclusion holds only in the normalized sense:
\begin{equation}
\liminf_{t\to\infty}
\mathbb{E}\!\left[
\frac{\|R_{t+1}-R_t\|_F^2}{\alpha_t^2}
\right]
\ge
\frac{(1-\kappa)\sigma^2}
{1+(\overline{\alpha}G/2)^2}.
\label{eq:normalized-displacement}
\end{equation}
\end{proposition}
The proof is presented in Appendix~\ref{app:a.2}.

Proposition~\ref{prop:persistent-ste-cayley} shows that once such a persistent STE residual exists and is not
systematically cancelled by the reference tangent direction, finite-step
Cayley updates transfer this residual into persistent motion on the
orthogonal manifold. This interpretation is consistent with prior analyses
of STE, which view the STE direction as a surrogate or coarse gradient whose
descent behavior depends on its alignment with an appropriate reference
gradient~\citep{bengio2013estimating,yin2019understanding}. It is also
consistent with Cayley-type optimization on the Stiefel manifold, where
standard convergence guarantees rely on smoothness and controlled gradient
behavior~\citep{liefficient}. Under non-decaying step sizes, the residual
floor leads to a non-vanishing displacement floor; under decaying learning
rates, only the normalized displacement in \eqref{eq:normalized-displacement}
can be claimed.


\section{Method}

Following the above analysis, we reveal fundamental convergence issues in quantized model optimization. The direct consequence of these issues is that even with substantial optimization costs, the resulting quantized models remain suboptimal. Therefore, we aim to employ a single rotation without gradient optimization to smooth outliers, for which we design two components, both composed of Givens orthogonal rotations.

\subsection{Givens Rotation}

In numerical linear algebra, the Givens rotation \citep{press2007numerical} represents a selective rotation within a two-dimensional plane. Formally, given a Givens orthogonal rotation denoted as $\mathbf{G}(i,j;\theta)$, the elements $cos(\theta)$  and $sin(\theta)$ occupy the intersection of rows $i$ and $j$ and columns $i$ and $j$ in $\mathbf{G}$, while the remaining non-zero elements lie along the diagonal. Geometrically, for a vector$x\in \mathbb{R}^d$, $x\mathbf{G}(i,j;\theta)$ signifies that vector $x$ is rotated clockwise by angle $\theta$ in the subspace plane spanned by the $i$-th and $j$-th coordinate axes.

Due to the limitations of quantization space, previous works \citep{liu2024spinquant,lin2024duquant} primarily focus on smoothing MO. Specifically, it can be expressed as:
\begin{equation}
	\label{xw}
	\mathbf{X}\mathbf{W}=(\mathbf{X}\mathbf{R})(\mathbf{R}^T\mathbf{W}) ,
\end{equation}
where the input activations are $\mathbf{X}\in \mathbb{R}^{T\times n}$, the corresponding weights are $\mathbf{W}\in \mathbb{R}^{n\times C_{out}}$, and $\mathbf{R}$ matrices are designed algorithmically. However, owing to the characteristic of MO being sparse yet exhibiting large magnitudes, Givens rotation matrices can specifically target and smooth the MO.

\begin{lemma}
\label{lem:givens-equalization}
Let $V=(a,b)\in\mathbb{R}^{1\times 2}$ and $r=\sqrt{a^2+b^2}$. Then
\begin{equation}
\min_{G\in O(2)}\|VG\|_\infty=\frac{r}{\sqrt{2}}.
\end{equation}
For $r>0$, this value is attained by the Givens rotation
\begin{equation}
G(\theta^\star)=
\begin{pmatrix}
\cos\theta^\star & -\sin\theta^\star\\
\sin\theta^\star & \cos\theta^\star
\end{pmatrix},
\quad
\theta^\star=\operatorname{atan2}(b,a)-\frac{\pi}{4},
\end{equation}
for which
\begin{equation}
VG(\theta^\star)=
\left(\frac{r}{\sqrt{2}},\frac{r}{\sqrt{2}}\right).
\end{equation}
\end{lemma}
The proof is presented in Appendix~\ref{app:a.3}.

Lemma \ref{lem:givens-equalization} demonstrates that by specifying rotation axes $(i,j)$ and angle $\theta^*$, the target vector is rotated within the two-dimensional subspace spanned by the $i$-th and $j$-th coordinate axes. This rotation balances the energy of the two rotated components, achieving smoothing of MO along specific components.

\subsection{SingleQuant}\label{singlequant method}
In this subsection, we introduce SingleQuant. This approach comprises two rotation components: For sparsely MO, we design an Alignment Rotation Transformation (ART) that precisely mitigates these outliers; For densely NO, we devise a Uniformity Rotation Transformation (URT) that adjusts their distribution through uniform mapping. Remarkably, SingleQuant requires only a a single calibration forward pass to complete quantization without any additional optimization.

\textbf{Kronecker Product.}
SingleQuant constructs a large rotation matrix 
$\mathbf{R}\in\mathbb{R}^{n\times n}$ from two smaller orthogonal matrices
$\mathbf{R}_1\in\mathbb{R}^{n_1\times n_1}$ and
$\mathbf{R}_2\in\mathbb{R}^{n_2\times n_2}$, where $n=n_1n_2$:
\begin{equation}
    \mathbf{R}=\mathbf{R}_1\otimes\mathbf{R}_2 .
    \label{eq:R-R1-R2}
\end{equation}

We use row-major vectorization. For 
$\mathbf{A}\in\mathbb{R}^{n_1\times n_2}$, define
$\operatorname{rvec}(\mathbf{A})\in\mathbb{R}^{1\times n}$ as the row-wise
flattening of $\mathbf{A}$. For a row vector $V\in\mathbb{R}^{1\times n}$,
let $V=\operatorname{rvec}(V_{\mathrm{mat}})$, where
$V_{\mathrm{mat}}\in\mathbb{R}^{n_1\times n_2}$. Then
\begin{equation}
    \label{eq18}
    V(\mathbf{R}_1\otimes\mathbf{R}_2)
    =
    \operatorname{rvec}
    \left(
    \mathbf{R}_1^T V_{\mathrm{mat}}\mathbf{R}_2
    \right).
\end{equation}
The Kronecker form further improves the efficiency of applying the rotation.
A direct multiplication with a dense matrix $\mathbf{R}\in\mathbb{R}^{n\times n}$
costs $\mathcal{O}(n^2)$ for each row vector. According to Eq.~(30), the
Kronecker-structured rotation can be applied by two smaller matrix
multiplications involving $\mathbf{R}_1\in\mathbb{R}^{n_1\times n_1}$ and
$\mathbf{R}_2\in\mathbb{R}^{n_2\times n_2}$, with total cost
$\mathcal{O}(n_1^2n_2+n_1n_2^2)$. Alg.~\ref{alg:factor-classic} chooses a balanced factorization
$n=n_1n_2$ with $n_1\approx n_2\approx \sqrt{n}$, reducing the rotation
application complexity from $\mathcal{O}(n^2)$ to $\mathcal{O}(n^{3/2})$.

Let $\tilde{\mathbf{X}}\in\mathbb{R}^{N\times n_1\times n_2}$ be obtained
by reshaping each row of $\mathbf{X}\in\mathbb{R}^{N\times n}$, and let
$\tilde{\mathbf{W}}\in\mathbb{R}^{C\times n_1\times n_2}$ be obtained by
reshaping each row of $\mathbf{W}^{\top}\in\mathbb{R}^{C\times n}$.
Let $\mathrm{Flat}(\cdot)$ denote the row-wise flattening operator, i.e.,
\begin{equation}
[\mathrm{Flat}(\tilde{\mathbf{X}})]_{t,:}
=
\mathrm{rvec}(\tilde{\mathbf{X}}_{t,:,:}),\quad
[\mathrm{Flat}(\tilde{\mathbf{W}})]_{c,:}
=
\mathrm{rvec}(\tilde{\mathbf{W}}_{c,:,:}).
\end{equation}

For a tensor $\mathcal{A}\in\mathbb{R}^{d_1\times d_2\times d_3}$, we use
the mode product convention
\begin{equation}
(\mathcal{A}\times_2\mathbf{M})_{i,p,j}
=
\sum_{q=1}^{d_2}\mathcal{A}_{i,q,j}\mathbf{M}_{p,q},\quad
(\mathcal{A}\times_3\mathbf{N})_{i,p,l}
=
\sum_{q=1}^{d_3}\mathcal{A}_{i,p,q}\mathbf{N}_{l,q}.
\end{equation}

Define
\begin{equation}
\tilde{\mathbf{X}}_R
=
\tilde{\mathbf{X}}\times_2\mathbf{R}_1^{\top}\times_3\mathbf{R}_2^{\top},
\qquad
\tilde{\mathbf{W}}_R
=
\tilde{\mathbf{W}}\times_2\mathbf{R}_1^{\top}\times_3\mathbf{R}_2^{\top}.
\end{equation}

Under the above convention, each slice satisfies
\begin{equation}
(\tilde{\mathbf{X}}_R)_{t,:,:}
=
\mathbf{R}_1^{\top}\tilde{\mathbf{X}}_{t,:,:}\mathbf{R}_2,
\qquad
(\tilde{\mathbf{W}}_R)_{c,:,:}
=
\mathbf{R}_1^{\top}\tilde{\mathbf{W}}_{c,:,:}\mathbf{R}_2.
\end{equation}

Therefore,
\begin{equation}
\mathrm{Flat}(\tilde{\mathbf{X}}_R)=\mathbf{X}\mathbf{R},
\qquad
\mathrm{Flat}(\tilde{\mathbf{W}}_R)=\mathbf{W}^{\top}\mathbf{R}
=
(\mathbf{R}^{\top}\mathbf{W})^{\top}.
\end{equation}

Since $\mathbf{R}$ is orthogonal, we obtain
\begin{equation}
\mathbf{X}\mathbf{W}
=
(\mathbf{X}\mathbf{R})(\mathbf{R}^{\top}\mathbf{W})
=
\mathrm{Flat}(\tilde{\mathbf{X}}_R)
\mathrm{Flat}(\tilde{\mathbf{W}}_R)^{\top}.
\end{equation}

In SingleQuant, $\mathbf{R}_1$ and $\mathbf{R}_2$ are both composed of ART and URT. Next, we introduce how to construct these two rotations.

\textbf{Alignment Rotation Transformation.} Conventional global rotation or progressive iterative methods \citep{shaoomniquant,liu2024spinquant,lin2024duquant} often struggle to simultaneously achieve precision localization and efficient suppression of outliers: the former lacks targeted handling of MO, while the latter introduces additional overhead. Directly smoothing outliers through a single rotational transformation presents significant challenges. Therefore, we design the ART. ART leverages the conclusion of Lemma \ref{lem:givens-equalization}, combining rapid MO detection with a single optimal rotation to smooth the most salient abnormal activations in a single transformation.

The construction expression for the ART matrix $\mathbf{R}^{A}\in \mathbb{R}^{n\times n}$ is as follows:
\begin{equation}
	\mathbf{R}^{A}=\begin{pmatrix} \mathbf{G}(\theta^*) & 0 \\ 0 & \mathbf{O} \end{pmatrix}\mathbf{P}_{ij} ,
\end{equation}
where $i$ and $j$ are the dimensions where the outlier and minimum value of the current activation $\mathbf{X}$ reside, $\mathbf{P}_{ij}$ is a permutation matrix that locates the outlier into the subspace $\mathbb{R}^{2\times 2}$. $\mathbf{O}\in \mathbb{R}^{(n-2)\times (n-2)}$ is a randomly orthogonalized matrix that preserves metric invariance in high-dimensional subspaces, preventing additional noise from rotational propagation while ensuring Givens rotation acts solely on target dimensions. 

ART smooths the selected outlier by computing the closed-form rotation angle
$\theta^{*}$ derived in Lemma~\ref{lem:givens-equalization}. Unlike learnable rotation methods that rely on
STE-based gradients over the orthogonal manifold, ART constructs the rotation
directly from activation statistics and does not backpropagate through the
quantizer. Therefore, it bypasses the instability mechanisms discussed in
Section~3.2, including STE-induced discontinuities near quantization boundaries
and finite-step Cayley update noise. The ablation results further verify that this closed-form smoothing step improves low-bit quantization performance.

\begin{algorithm}[thb]
	\begin{algorithmic}[1]
		\caption{Kronecker Dimension Factorization}\label{alg:factor-classic}
		\REQUIRE Hidden layer dimension $n$ ($n \geq 1$)
		\ENSURE Factor pair $(n_1, n_2)$ satisfying $n = n_1 \times n_2$ and $n_1 \approx \sqrt{n}$
		\STATE $\sqrt{n} \gets \sqrt{n}$
		\STATE $n_2 \gets 1$
		\STATE $k_{\max} \gets \lfloor \log_2(n) \rfloor$
		\FOR{$k = 0$ \TO $k_{\max}$}
		\STATE $a \gets 2^k$ \COMMENT{Compute $2^k$}
		\IF{$n \bmod a \neq 0$}
		\STATE \textbf{CONTINUE}
		\ENDIF
		\IF{$|a - \sqrt{n}| < |n_2 - \sqrt{n}|$}
		\STATE $n_2 \gets a$ \COMMENT{Update the optimal factor}
		\ENDIF
		\ENDFOR
		\STATE $n_1 \gets n / n_2$
		\STATE \RETURN{$(n_1, n_2)$}
	\end{algorithmic}
\end{algorithm}

\textbf{Uniformity Rotation Transformation.} Although ART smooths MO, a substantial number of NO \citep{lin2024duquant} persist in LLM activations. These outliers exhibit consistent median values across specific feature dimensions and exist in all token sequences \citep{xiao2023smoothquant}. We design the URT, which constructs a norm-preserving rotation toward a rank-preserving uniform target. Specifically, for a row vector $V$, we aim to construct a rotation matrix $\mathbf{R}^{U}$ such that:
\begin{equation}
    V \mathbf{R}^{U}=U\in \mathbb{R}^{n},
\end{equation}
where $U$ is a norm-preserving uniform target vector constructed from $V$. Let $\pi$ sort the entries of $V$ such that
\begin{equation}
    V_{\pi(1)} \leq \cdots \leq V_{\pi(n)} .
\end{equation}
We define a centered uniform template $q\in\mathbb{R}^{n}$ as
\begin{equation}
    q_k=\frac{2k-n-1}{n}, \qquad k=1,\ldots,n ,
\end{equation}
and assign it to $U$ according to the rank order of $V$:
\begin{equation}
    U_{\pi(k)}=\frac{\|V\|_F}{\|q\|_F}q_k, \qquad k=1,\ldots,n .
\end{equation}
Thus, $U$ preserves the relative ordering of $V$ while satisfying $\|U\|_F=\|V\|_F$, and its entries are evenly spread over a centered uniform range. Next, we explain how to construct $\mathbf{R}^{U}$. By respectively mapping $V$ and $U$, we obtain:
\begin{equation}
	V\mathbf{R}_{map}=U\mathbf{R}'_{map}=\|V \|_F  e_1^{T} ,
\end{equation}
where $e_1$is the standard basis vector. Since Ma et al. demonstrated that two vectors with identical norms can be mutually transformed by $n-1$ Givens rotations \citep{ma2024parameter}, thus through $O(n)$ complexity mapping of $U$ and $V$ onto $e_1^{T}$, we obtain $\mathbf{R}_{map}$ and $\mathbf{R}'_{map}$. Thus the uniformity rotation matrix $\mathbf{R}^{U}$ can be expressed as:
\begin{equation}
	\mathbf{R}^{U}=\mathbf{R}_{map}(\mathbf{R}'_{map})^{T} .
\end{equation}

Similar to the ART, the URT is entirely based on closed-form rotation matrix construction without additional gradient calculation or iterative tuning. Moreover, $\mathbf{R}^{U}$ flattens the original activations into a uniform distribution and reduces quantization noise shift caused by outliers.

\textbf{The Overall SingleQuant Method.} To simultaneously address the smoothing of MO and NO, we first employ ART to smooth MO, eliminating quantization obstacles caused by extreme values. Next, we introduce the URT to flatten and uniformly distribute activations for quantization adaptation. Overall, $\mathbf{R}$ in Equation \eqref{eq:R-R1-R2} can be expressed as:
\begin{equation}
	\mathbf{R}=(\mathbf{R}^{U}_1\mathbf{R}^{A})^{T}\otimes (\mathbf{H}\mathbf{R}^{U}_2) ,
\end{equation}
where $\mathbf{H}$ denotes the Hadamard matrix that shares the same shape as $\mathbf{R}^{U}_2$, and $\mathbf{R}^{U}_1$ and $\mathbf{R}^{U}_2$ represent uniformity rotation matrices built according to different dimensions of the activation $\mathbf{X}$.

In summary, SingleQuant employs a unified transformation framework to seamlessly integrate ART and URT, achieving synchronous smoothing of both activations and weights. Notably, for weight processing, owing to the orthogonal property of rotation matrices, we apply structurally equivalent rotations to the original weight matrix. Formally, the SingleQuant methodology not only simultaneously mitigates long-tail risks in the distributions of both activations and weights, significantly reducing precision degradation caused by extreme outliers, but also achieves low-overhead transformation integration through rotation fusion techniques.

\begin{table*}[tb]
	\centering
	\caption{WikiText-2 and C4 perplexity of 4-bit weight \& activation quantized LLaMA models. $\downarrow$ denotes that the smaller the score, the better the performance. Bold values denote the best performance scores.}
	\begin{adjustbox}{width=\textwidth, center} 
		\begin{tabular}{l c c c c c c c c c c c}
			\toprule
			\multirow{2.4}{*}{Method} & \multirow{2.4}{*}{W Quant.} & \multicolumn{5}{c}{WikiText-2 \textdownarrow} & \multicolumn{5}{c}{C4 \textdownarrow} \\
			\cmidrule(lr){3-7} \cmidrule(lr){8-12}
			& & 2-7B & 2-13B & 2-70B & 3-8B & 3-70B & 2-7B & 2-13B & 2-70B & 3-8B & 3-70B \\
			\midrule
			FP16 & - & 5.47 & 4.88 & 3.32 & 6.14 & 2.86 & 7.26 & 6.73 & 5.71 & 9.45 & 7.17 \\
			SmoothQuant & RTN & 83.12 & 35.88 & 26.01 & 210.19 & 9.60 & 77.27 & 43.19 & 34.61 & 187.93 & 16.90 \\
			OmniQuant & RTN & 14.74 & 12.28 & - & - & - & 21.40 & 16.24 & - & - & - \\
			AffineQuant & RTN & 12.69 & 11.45 & - & - & - & 15.76 & 13.97 & - & - & - \\
			QuaRot & RTN & 8.56 & 6.10 & 4.14 & 10.60 & 55.44 & 11.86 & 8.67 & 6.42 & 17.19 & 79.48 \\
			QuaRot & GPTQ & 6.10 & 5.40 & 3.79 & 8.16 & 6.60 & 8.32 & 7.54 & 6.12 & 13.38 & 12.87 \\
			
			QUIK-4B & GPTQ & 8.87 & 7.78 & 6.91 & - & - & - & - & - & - & - \\
			SpinQuant & RTN & 6.14 & 5.44 & 3.82 & 7.96 & 7.58 & 9.19 & 8.11 & 6.26 & 13.45 & 15.39 \\
			SpinQuant & GPTQ & \textbf{5.96} & 5.24 & 3.70 & \textbf{7.39} & 6.21 & 8.28 & 7.48 & 6.07 & 12.19 & 12.82 \\
            MergeQuant&GPTQ&6.09&5.29&3.78&7.92&6.86&7.87&6.98&5.89&11.71&10.52\\
			DuQuant & RTN & 6.28 & 5.42 & 3.79 & 8.56 & 6.06 & 7.90 & 7.05 & 5.87 & 11.98 &9.63  \\ 
            \midrule
			 \rowcolor{pink!20}{SingleQuant} (Ours) & RTN & 6.12\tiny{$\pm$0.03} & \textbf{5.22}\tiny{$\pm$0.02} & \textbf{3.65}\tiny{$\pm$0.04} & 7.86\tiny{$\pm$0.07} & \textbf{4.71} \tiny{$\pm$0.05}& \textbf{7.60}\tiny{$\pm$0.12} & \textbf{6.82} \tiny{$\pm$0.14}& \textbf{5.80}\tiny{$\pm$0.09} & \textbf{11.36}\tiny{$\pm$0.12} &\textbf{8.01}\tiny{$\pm$0.11} \\
			\bottomrule
		\end{tabular}
	\end{adjustbox}
    
	\label{tab1}
\end{table*}

\section{Experiments}

\subsection{Experimental Settings}

\textbf{Evaluation Benchmarks.}
We evaluate SingleQuant on a diverse set of representative LLM architectures, including LLaMA-2 \citep{touvron2023llama}, LLaMA-3 \citep{grattafiori2024llama}, Vicuna-v1.5 \citep{chiang2023vicuna}, and Mixtral 8$\times$7B. The evaluation covers three categories of downstream tasks. First, for language generation, we report perplexity on WikiText-2 \citep{merity2017pointer} and C4 \citep{raffel2020exploring}, which are widely used to measure the modeling quality of quantized LLMs. Second, for zero-shot reasoning and question answering, we evaluate six standard benchmarks, including ARC-Challenge and ARC-Easy \citep{clark2018think}, HellaSwag \citep{zellers2019hellaswag}, LAMBADA \citep{paperno2016lambada}, PIQA \citep{bisk2020piqa}, and WinoGrande \citep{sakaguchi2021winogrande}. Third, to assess broad knowledge and multi-task understanding, we evaluate the quantized models on MMLU \citep{hendrycksmeasuring} under both zero-shot and five-shot settings when applicable. We mainly focus on the W4A4 setting, where both weights and activations are quantized to 4 bits. This setting is substantially more challenging than weight-only quantization due to the severe outlier sensitivity of low-bit activation quantization, while being practically relevant for memory- and bandwidth-efficient LLM inference \citep{xiao2023smoothquant,ashkboos2024quarot,liu2024spinquant}.

\textbf{Baselines.}
We compare SingleQuant with a broad range of post-training quantization methods under 4-bit weight-activation quantization settings. The baseline methods include SmoothQuant \citep{xiao2023smoothquant}, OmniQuant \citep{shaoomniquant}, QuaRot \citep{ashkboos2024quarot}, DuQuant \citep{lin2024duquant}, MergeQuant \citep{wang2025mergequant}, and other competitive INT4 quantization approaches. In particular, we include two recent rotation-based quantization methods, SpinQuant \citep{liu2024spinquant} and OSTQuant \citep{huostquant}, as the main strong baselines. For methods supporting different weight quantizers, we report results with RTN and GPTQ \citep{frantaroptq} following their original settings, enabling a fair comparison between optimization-based, rotation-based, and our closed-form rotation construction.

\textbf{Implementation Details.} We implement the quantization and evaluation of SingleQuant building upon the Huggingface \citep{wolf2019huggingface}, PyTorch \citep{paszke2019pytorch}, and lm-evaluation-harness \citep{eval-harness} frameworks. To ensure experimental accuracy and mitigate randomness, each SingleQuant experiment was executed across 10 distinct random seeds, with the results reported representing the averaged metrics.

\subsection{Main Results}

\textbf{Results on Language Generation Tasks.} Tab.~\ref{tab1} reports SingleQuant's perplexity results on WikiText-2 and C4 datasets. Notably, SingleQuant with RTN weight quantizer consistently outperforms prior quantization methods across most benchmarks, exhibiting particularly superior performance on larger models. Despite significant degradation in low-bit quantization for LLaMA-3 \citep{huang2024good}, on LLaMA-3-70B/C4, SingleQuant is only 0.84 perplexity points higher than FP16. Remarkably, RTN-based SingleQuant is competitive with or surpasses most GPTQ-based alternatives in several settings, demonstrating that our ART/URT components effectively manage outliers and compensate for RTN's limitations.

\textbf{Results on Zero-shot QA Tasks.} On QA tasks, we evaluate six zero-shot tasks as shown in Table \ref{tab2}. SingleQuant further narrows the performance gap between the quantized model and the FP16 baseline. For instance, on the LLaMA-2-70B model, SingleQuant is only 0.91\% lower in average accuracy than FP16, while other baseline methods suffer greater losses exceeding 1\%. More detailed results are reported in Appendix~Tab.~\ref{app:tab2}. These substantial improvements demonstrate that across diverse tasks, the two proposed components exhibit significant efficacy in smoothing weight and activation outliers.

\textbf{Results on Vicuna-MMLU Tasks.} To further examine the generalization of SingleQuant beyond base LLaMA models, we evaluate Vicuna-v1.5 on MMLU under both zero-shot and five-shot settings, as shown in Tab.~\ref{vicuna-mmlu}. SingleQuant consistently achieves the best performance among quantized methods. On Vicuna-v1.5-7B, SingleQuant obtains 44.30\% and 45.25\% average accuracy under zero-shot and five-shot settings, respectively, outperforming DuQuant by 0.65\% and 1.62\%. On Vicuna-v1.5-13B, SingleQuant further improves the five-shot accuracy to 52.22\%, reducing the gap to the FP16 model. These results indicate that the proposed rotation construction remains effective for instruction-tuned models and preserves knowledge-intensive reasoning ability under W4A4 quantization.

\textbf{Results on Mixtral Models.} We also evaluate SingleQuant on Mixtral 8$\times$7B to verify its applicability to MoE architectures. As reported in Tab.~\ref{moe}, SingleQuant achieves the lowest perplexity among all quantized baselines, with 4.38 on WikiText-2 and 7.25 on C4. Compared with DuQuant, SingleQuant reduces perplexity by 0.13 and 0.34 on the two datasets, respectively. This demonstrates that SingleQuant is not limited to dense Transformer models; its closed-form rotation design can also handle the heterogeneous activation distributions introduced by expert routing in MoE models.

\begin{table}[htb]
	\centering
	\caption{Zero-shot\textsuperscript{6} AVG. results of 4-bit weight \& activation quantized LLaMA models. * denotes the usage of GPTQ weight quantization. The official OSTQuant repository lacks support for parallel training and inevitably encounters out-of-memory (OOM) errors when processing 70B models.}
	\begin{tabular}{l c c c c c }
		\toprule
		\multirow{2.4}{*}{Method}  & \multicolumn{5}{c}{Zero-shot\textsuperscript{6} AVG. \textuparrow}\\
		\cmidrule(lr){2-6}      
		& 2-7B & 2-13B & 2-70B & 3-8B & 3-70B  \\
		\midrule
		FP16&69.87&72.55&77.05&73.23&79.95\\	
		QuaRot&57.73&66.25&73.47&61.34&35.36\\	
		QuaRot\textsuperscript{*}&65.01&68.91&75.68&65.79		&65.37\\	
		SpinQuant&63.52&68.56&75.09&66.98&65.66\\	
		SpinQuant\textsuperscript{*}&66.23&70.93&76.06&	68.70	&71.33\\	
		DuQuant&61.34&64.98&69.39&65.76&72.97\\	
        OSTQuant\textsuperscript{*}&66.39&70.27&OOM&\textbf{69.08}&OOM\\
        \midrule
		 \rowcolor{pink!20}{SingleQuant}&\textbf{67.18}&\textbf{71.50}&\textbf{76.14}&68.96	&\textbf{76.30}\\	
		\bottomrule
	\end{tabular}
	\label{tab2}
\end{table}

\begin{table*}[ht]
	\centering
	\caption{Zero-shot and five-shot results on the MMLU benchmark for Vicuna-v1.5 models under 4-bit weight-activation quantization.}
	\label{vicuna-mmlu}
	\begin{adjustbox}{width=\linewidth,center}
		\begin{tabular}{l|l|c c c c c|c c c c c}
			\toprule
			\multirow{2}{*}{Model} & \multirow{2}{*}{Method} 
			& \multicolumn{5}{c|}{MMLU (0 shot) ↑} 
			& \multicolumn{5}{c}{MMLU (5 shot) ↑} \\
			\cmidrule{3-7} \cmidrule{8-12}
			& & STEM & Hums & Social & Others & Avg. 
			& STEM & Hums & Social & Others & Avg. \\
			\midrule
			\multirow{6}{*}{Vicuna-v1.5-7B} 
			& FP16 & 38.67 & 45.31 & 56.19 & 56.05 & 48.75 & 39.63 & 45.76 & 58.14 & 57.43 & 49.85 \\
			& SmoothQuant & 27.10 & 25.16 & 27.40 & 26.71 & 26.59 & 25.22 & 25.06 & 24.99 & 26.68 & 25.49 \\
			& OmniQuant & 27.20 & 24.00 & 27.14 & 25.08 & 25.86 & 29.39 & 24.95 & 27.30 & 24.80 & 26.39 \\
			& Atom & 30.28 & 34.73 & 38.97 & 40.56 & 36.14 & 31.97 & 35.37 & 40.46 & 40.81 & 37.15 \\
			& DuQuant & \textbf{35.49} & 41.32 & 49.11 & 49.48 & 43.65 & 36.08 & 40.38 & 49.14 & 50.12 & 43.63 \\
			& \pinkcell{SingleQuant }
			&\pinkcell{ 34.79} & \pinkcell{\textbf{41.91}} & \pinkcell{\textbf{50.83}} & \pinkcell{\textbf{50.43}} & \pinkcell{\textbf{44.30}} 
			& \pinkcell{\textbf{36.78}} & \pinkcell{\textbf{41.87}} & \pinkcell{\textbf{51.84}} & \pinkcell{\textbf{51.79}} & \pinkcell{\textbf{45.25}} \\
			\midrule
			\multirow{6}{*}{Vicuna-v1.5-13B} 
			& FP16 & 43.77 & 50.50 & 62.69 & 62.74 & 54.55 & 44.83 & 51.94 & 65.23 & 62.37 & 55.73 \\
			& SmoothQuant & 21.70 & 24.29 & 22.13 & 23.16 & 22.82 & 25.31 & 24.97 & 26.00 & 27.08 & 25.84 \\
			& OmniQuant & 26.81 & 26.57 & 30.35 & 28.75 & 28.12 & 28.79 & 27.29 & 31.13 & 28.99 & 29.05 \\
			& Atom & 32.54 & 39.60 & 46.02 & 46.11 & 41.07 & 35.35 & 39.21 & 59.72 & 45.77 & 45.01 \\
			& DuQuant & 40.62 & 48.03 & 59.05 & 58.20 & 51.20 & 41.92 & 48.65 & 58.82 & 57.56 & 51.49 \\
			& \pinkcell{SingleQuant }
			& \pinkcell{\textbf{40.97}} & \pinkcell{\textbf{48.13}} & \pinkcell{\textbf{59.06}} & \pinkcell{\textbf{58.69}} & \pinkcell{\textbf{51.42}}
			& \pinkcell{\textbf{42.04}} & \pinkcell{\textbf{49.14}} & \pinkcell{\textbf{59.87}} & \pinkcell{\textbf{58.91}} & \pinkcell{\textbf{52.22}} \\
			\bottomrule
		\end{tabular}
	\end{adjustbox}
\end{table*}

\begin{table}[h]
	\centering
	\caption{The comparison of perplexity for the two datasets under 4-bit weight-activation quantization, with the model using the MoE model Mixtral 8$\times$7B.}
\renewcommand\tabcolsep{10.5pt}
	\begin{tabular}{llcc}
		\toprule
		LLM & Method & Wikitext \ \textdownarrow & C4 \ \textdownarrow \\
		\midrule
		\multirow{6}{*}{\makecell[c]{Mixtral\\8$\times$7B}} & FP16  & 3.84 & 6.91 \\
		& QuaRot  & 9.06  & 12.91 \\
		& AWQ  & 5.96 & 8.77 \\
		& DuQuant & 4.51 & 7.59 \\
		& \pinkcell{SingleQuant} & \pinkcell{\textbf{4.38}}  & \pinkcell{\textbf{7.25}} \\
		\bottomrule
	\end{tabular}
	\label{moe}
\end{table}

\subsection{Comparison with FlatQuant}
Here, we specifically highlight the comparison with FlatQuant, which also utilizes the Kronecker product to construct rotation matrices. Notably, FlatQuant incorporates Learnable Clipping Thresholds (LCT); therefore, we report the comparison between SingleQuant and FlatQuant under equivalent settings in Table \ref{tab:with flat}. As observed, SingleQuant outperforms FlatQuant under the same conditions, validating the effectiveness of our proposed ART and URT. Furthermore, this demonstrates that these two types of rotations and the Kronecker product are complementary, achieving superior smoothing effects with $O(n^{3/2})$ efficiency.

\begin{table}[t]
\centering
\renewcommand{\arraystretch}{1.2}
\caption{Comparison results of SingleQuant and FlatQuant. \textdagger \ denotes data from the original FlatQuant paper. PPL AVG. denotes mean WikiText-2 and C4 perplexity; 0-shot\textsuperscript{6} AVG. denotes average across 6 zero-shot tasks.}
\label{tab:with flat}
\begin{tabular}{llcccc}
\toprule
\multicolumn{2}{c}{\multirow[c]{2}{*}[-1.4ex]{\textbf{Configuration}}} & \multicolumn{2}{c}{2-13B} & \multicolumn{2}{c}{3-8B} \\
\cmidrule(lr){3-4} \cmidrule(lr){5-6}
\multicolumn{2}{c}{} &\makecell[c]{{PPL}\\{AVG.$\downarrow$}} &\makecell[c]{{0-shot\textsuperscript{6}}\\{AVG.$\uparrow$}} & \makecell[c]{{PPL}\\{AVG.$\downarrow$}} & \makecell[c]{{0-shot\textsuperscript{6}}\\{AVG.$\uparrow$}} \\
\midrule

\multirow{2}{*}{w/ LCT} & Flat.   & 6.11 & 71.64 & 9.06\textsuperscript{\textdagger} & 71.33\textsuperscript{\textdagger} \\
                        & Single. & \textbf{5.76} & \textbf{72.19} & \textbf{9.03} & \textbf{71.91} \\ 
\cmidrule(lr){1-6} 

\multirow{2}{*}{w/o LCT}  & Flat.   & 7.14 & 70.17 & 10.35\textsuperscript{\textdagger} & 67.08\textsuperscript{\textdagger} \\
& Single. & \textbf{6.02} & \textbf{71.50} &\textbf{9.61} & \textbf{68.96} \\
\bottomrule
\end{tabular}
\end{table}

\subsection{Ablation Study}

\begin{table}[htb]
	\centering
	\caption{Effect of component ablation on LLaMA-2/3: PPL AVG. denotes mean WikiText-2 and C4 perplexity; 0-shot\textsuperscript{6} AVG. denotes average across 6 zero-shot tasks.}
	\label{ablation}
\renewcommand\tabcolsep{8pt}
	\begin{tabular}{@{}cccccc@{}}
		\toprule
		\multicolumn{2}{c}{{Rotation}} & 
		\multicolumn{2}{c}{{2-13B}} & 
		\multicolumn{2}{c}{{3-8B}} \\
		\cmidrule(r){1-2} \cmidrule(lr){3-4} \cmidrule(l){5-6}
		\makecell[c]{{ART}\\(${R}^{A}$)} & \makecell[c]{{URT}\\(${R}^{U}$)} & 
		\makecell[c]{{PPL}\\{AVG.$\downarrow$}} & \makecell[c]{{0-shot\textsuperscript{6}}\\{AVG.$\uparrow$}} & 
		\makecell[c]{{PPL}\\{AVG.$\downarrow$}} & \makecell[c]{{0-shot\textsuperscript{6}}\\{AVG.$\uparrow$}} \\
		\midrule
		& &7.15 &68.91 & 11.82& 63.44 \\
		& \checkmark & 6.89 & 69.13 & 10.52 &64.17 \\
		\checkmark & & 6.10 &71.23 & 9.93 & 68.24 \\
		\checkmark & \checkmark & 6.02 & 71.50 & 9.74 & 68.96 \\
		\bottomrule
	\end{tabular}

\end{table}

\textbf{Influence of ART/URT.}
We conducted an ablation study on SingleQuant, focusing on the independent contributions and synergistic effects of the proposed ART and URT components. Experiments on LLaMA-2 and LLaMA-3 (Table \ref{ablation}) demonstrate that: ART significantly enhances quantized model accuracy, validating its efficacy in smoothing MO to preserve performance—consistent with prior conclusions \citep{jinmassive}; furthermore, while URT alone yields limited gains, its integration with ART produces marked synergy. We attribute this to complementary mechanisms: ART reduces MO magnitudes, increasing NO prevalence with non-uniform distributions, while URT remaps these outliers into flatter distributions, optimizing quantization space utilization. These findings corroborate Fig.~\ref{overall}c: ART resolves MO magnitude issues and URT adjusts NO distribution structures, jointly enabling comprehensive outlier processing.
\begin{figure}
	\centering
	\includegraphics[width=0.5\textwidth]{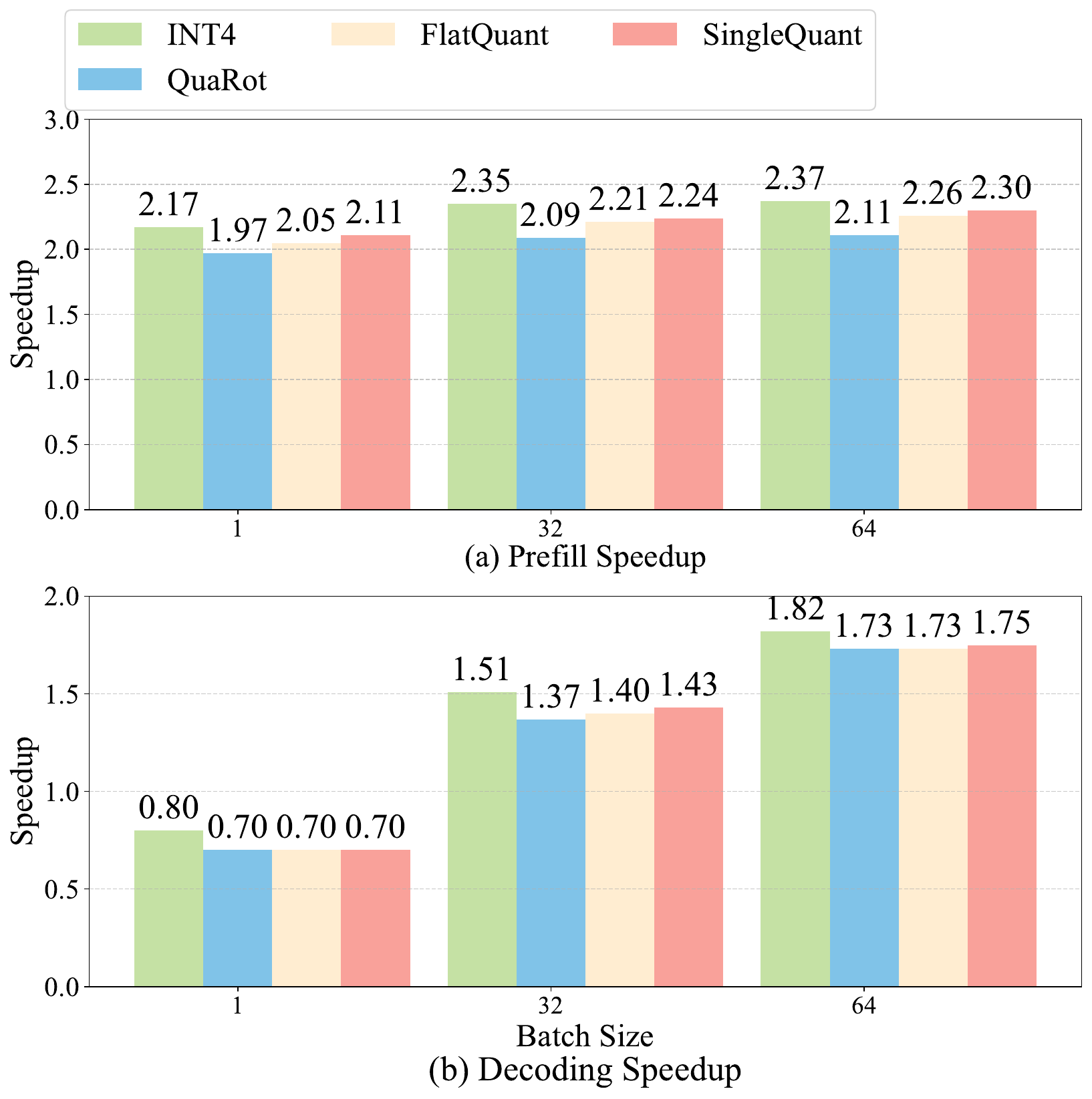} 
	\caption{Prefill and decoding speedup of LLaMA-2-7B model across different batch sizes. We decode 256 tokens after the prefill on a sequence length of 2048.}
	\label{speedup}
\end{figure}

\begin{figure}
\centerline{\includegraphics[width=0.5\textwidth]{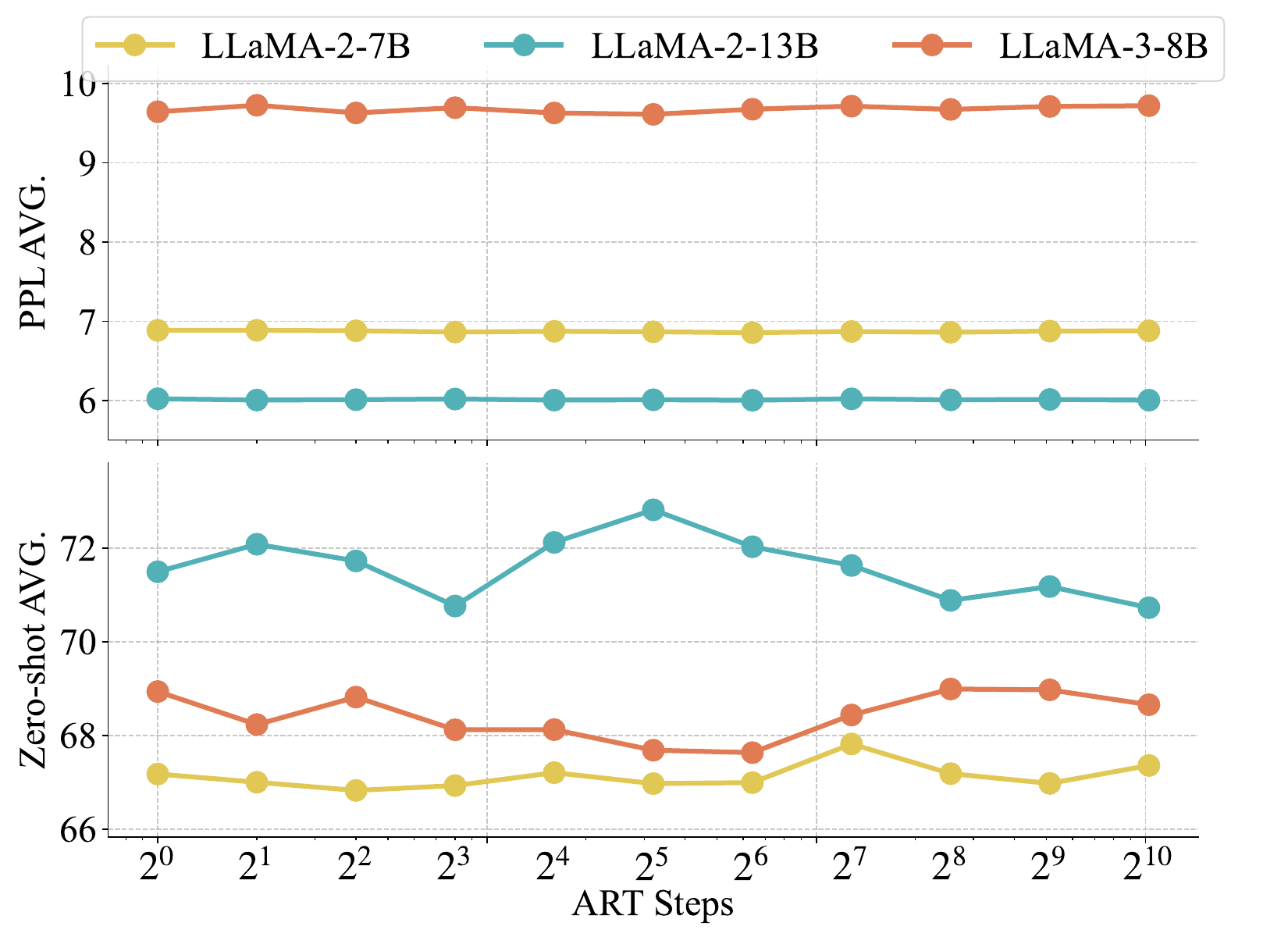}}
\caption{Performance of SingleQuant with varying ART steps. PPL AVG. denotes the mean
perplexity on WikiText-2 and C4, and Zero-shot AVG. denotes the mean accuracy on
ARC-C, ARC-E, HellaSwag, LAMBADA, PIQA, and WinoGrande.}
\label{singlequant optimizaiton}
\end{figure}

\textbf{Effect of ART Steps.} We further examine whether repeated ART rotations
are necessary for smoothing MO. As shown in Fig.~\ref{singlequant optimizaiton}, increasing the
number of ART steps from $2^{0}$ to $2^{10}$ does not bring consistent gains in
either PPL AVG. or zero-shot AVG. across LLaMA-2-7B, LLaMA-2-13B, and
LLaMA-3-8B. The performance remains largely saturated after the first rotation,
with only minor fluctuations as more rotations are applied. This result verifies
that a single closed-form Givens rotation is sufficient to smooth the dominant
MO in the selected subspace. Therefore, ART does not require iterative
refinement, which supports the single-pass design of SingleQuant and helps
maintain its low quantization overhead.

\textbf{Quantization Time.}
Compared to existing baselines, SingleQuant demonstrates significant quantization speed advantages. Through the Kronecker decomposition in Equation~\eqref{eq18} ensuring $O(n^{3/2})$ complexity order, and ART/URT matrices are single-pass optimized, SingleQuant is substantially faster than gradient-optimized or greedy algorithms. As shown in Tab.~\ref{runtime}, SingleQuant quantizes a 13B model in 37 seconds, achieving up to $1420\times$ acceleration over baseline methods. 
\begin{table}[htbp]
	\centering
	\caption{
    Comparisons of quantization time cost by quantizing various LLMs. We run each experiment 10 times on a single NVIDIA A800 80GB GPU and report the averaged time.}
\renewcommand\tabcolsep{7.5pt}
	\begin{tabular}{lcccc}
		\toprule
		{LLM} &   \makecell[c]{{OST}\\ {Quant}} &\makecell[c]{{Spin}\\ {Quant}} &  \makecell[c]{{Single}\\ {Quant}} \\
		\midrule
		LLaMA2-7B & 2,184s    & 22,334s         & 24s   \\
		LLaMA2-13B & 4,598s & 52,561s & 37s   \\
		LLaMA3-8B & 2,226s  & 24,849s     &27s  \\
		\bottomrule
	\end{tabular}
    \label{runtime}

\end{table}

\textbf{Inference Efficiency.} To evaluate the inference speedup provided by SingleQuant, we adopted the measurement strategy and inference kernel from \citep{sunflatquant}. As shown in Fig.~\ref{speedup}, during the Prefill phase, SingleQuant achieves a 2.3$\times$ speedup at batch size 64, only 0.07$\times$ lower than INT4. Notably, its speedup exhibits strong robustness to batch size variations, significantly outperforming existing baselines across different batch scales. During the decoding phase, SingleQuant achieves a 1.43$\times$ speedup over FP16 at a batch size of 32, outperforming the FlatQuant baseline by 0.03$\times$. This is attributed to the efficient dimension decomposition (Alg.~\ref{alg:factor-classic}), which reduces matrix multiplication complexity to $O(n^{3/2})$. While a minor gap persists versus INT4, SingleQuant maintains significant advantages in preserving model accuracy, making it more suitable for practical deployment scenarios given its superior accuracy-performance trade-off.

\textbf{Memory Usage.}
Tab.~\ref{memory} reports the peak memory usage with batch size 1 during the
prefill and decoding stages. Compared with the FP16 model, SingleQuant reduces
the peak memory from 15.28 GB to 4.61 GB in prefill and from 13.63 GB to
3.76 GB in decoding, achieving $3.31\times$ and $3.63\times$ memory savings,
respectively. Moreover, SingleQuant shows slightly lower memory consumption
than existing W4A4 baselines such as SmoothQuant, QuaRot, and DuQuant in both
stages. This indicates that the proposed rotation construction does not
introduce additional memory burden; instead, it preserves the memory advantage
of low-bit quantization while maintaining better accuracy and inference
efficiency.

\begin{table}[htbp]
\centering
\caption{Peak memory usage with a batch size of 1.}
\label{memory}
\small
\setlength{\tabcolsep}{4pt}
\begin{tabular}{lcccc}
\toprule
\textbf{Method} & \textbf{Prefill} & \textbf{Saving} & \textbf{Decode} & \textbf{Saving} \\
 & \textbf{(GB)} &  & \textbf{(GB)} &  \\
\midrule
FP16         & 15.28 & -              & 13.63 & -              \\
SmoothQuant  & 4.78  & 3.19$\times$   & 3.89  & 3.50$\times$   \\
QLLM         & 5.34  & 2.85$\times$   & 3.89  & 3.50$\times$   \\
QuaRot       & 4.78  & 3.19$\times$   & 3.89  & 3.50$\times$   \\
DuQuant      & 4.78  & 3.19$\times$   & 3.89  & 3.50$\times$   \\
\pinkcell{SingleQuant}  & \pinkcell{4.61}  & \pinkcell{3.31$\times$ }  & \pinkcell{3.76}  & \pinkcell{3.63$\times$}   \\
\bottomrule
\label{memory}
\end{tabular}
\end{table}

\section{Conclusion}
In this paper, we propose SingleQuant, an optimization-free rotation framework for low-bit weight-activation quantization of large language models. We first analyze the instability of STE-based rotation learning and show that quantization-induced discontinuities can lead to non-smooth gradients and persistent Cayley update oscillations. Motivated by this observation, SingleQuant replaces iterative rotation optimization with closed-form Givens-rotation constructions. Specifically, ART directly smooths sparse massive outliers through locally optimal two-dimensional rotations, while URT further reshapes normal outliers toward a more uniform and quantization-friendly distribution. To reduce the computational cost of applying large orthogonal transformations, we introduce a Kronecker-structured rotation design, enabling efficient transformation with substantially lower complexity. Extensive experiments on multiple LLM architectures and downstream benchmarks demonstrate that SingleQuant achieves strong W4A4 quantization performance while significantly reducing quantization time and preserving inference efficiency. These results indicate that deterministic, mathematically constructed rotations provide an effective and practical alternative to gradient-based quantization optimization for efficient LLM deployment.

\appendix
\renewcommand{\thesection}{\Alph{section}}

\makeatletter
\@addtoreset{equation}{section}
\@addtoreset{figure}{section}
\@addtoreset{table}{section}
\makeatother

\renewcommand{\theequation}{\thesection.\arabic{equation}}
\renewcommand{\thefigure}{\thesection.\arabic{figure}}
\renewcommand{\thetable}{\thesection.\arabic{table}}
\section{Proof}
\subsection{Proof of Proposition 1}\label{app:a.1}
\begin{proof}
Let $\mathbf R(s)\subset\mathcal M$ be a smooth curve satisfying
\begin{equation}
\mathbf R(0)=\mathbf R_0,\quad\mathbf R'(0)=\boldsymbol\xi.
\end{equation}
Since $D Z_\rho(\mathbf R_0)[\boldsymbol\xi]\neq 0$, Taylor expansion gives
\begin{equation}
Z_\rho(\mathbf R(s))
=
Z_\rho(\mathbf R_0)
+
s\,D Z_\rho(\mathbf R_0)[\boldsymbol\xi]
+
o(s).
\end{equation}
Denote
\begin{equation}
\gamma:=D Z_\rho(\mathbf R_0)[\boldsymbol\xi]\neq 0.
\end{equation}
Then
\begin{equation}
Z_\rho(\mathbf R(s))
=
\tau+s\gamma+o(s).
\end{equation}
Thus, for sufficiently small $s>0$, the two values
$Z_\rho(\mathbf R(s))$ and $Z_\rho(\mathbf R(-s))$ lie on opposite sides of
the quantization boundary $\tau$.

Since $\tau\in\mathcal B_\Delta$, there exists $k\in\mathbb Z$ such that
\begin{equation}
\tau=\left(k+\frac{1}{2}\right)\Delta.
\end{equation}
Therefore,
\begin{equation}
\lim_{\varepsilon\to 0^+}Q_\Delta(\tau-\varepsilon)=k\Delta,
\quad
\lim_{\varepsilon\to 0^+}Q_\Delta(\tau+\varepsilon)=(k+1)\Delta.
\end{equation}
Hence the jump of the quantizer across the boundary is $\Delta$. Consequently,
the one-sided limits of the Riemannian STE gradient estimator satisfy
\begin{equation}
\lim_{s\to 0^+}
\mathbf G_{\mathcal M}^{\mathrm{STE}}(\mathbf R(s))
-
\lim_{s\to 0^+}
\mathbf G_{\mathcal M}^{\mathrm{STE}}(\mathbf R(-s))
=
\pm
\Delta
\Pi_{\mathbf R_0}
\left[
D\mathbf Z(\mathbf R_0)^*\mathbf e_\rho
\right].
\end{equation}
By the non-degeneracy assumption,
\begin{equation}
\mathbf v=
\Pi_{\mathbf R_0}
\left[
D\mathbf Z(\mathbf R_0)^*\mathbf e_\rho
\right]
\neq \mathbf 0.
\end{equation}
Therefore,
\begin{equation}
\lim_{s\to 0^+}
\left\|
\mathbf G_{\mathcal M}^{\mathrm{STE}}(\mathbf R(s))
-
\mathbf G_{\mathcal M}^{\mathrm{STE}}(\mathbf R(-s))
\right\|_F
=
\Delta\|\mathbf v\|_F>0.
\end{equation}
On the other hand, since $\mathbf R(s)$ is smooth,
\begin{equation}
\|\mathbf R(s)-\mathbf R(-s)\|_F
=
2s\|\boldsymbol\xi\|_F+o(s)
\rightarrow 0
\quad
\text{as }s\to 0^+.
\end{equation}
If $\mathbf G_{\mathcal M}^{\mathrm{STE}}$ were Lipschitz continuous in a
neighborhood of \(\mathbf R_0\), then there would exist $L>0$ such that
\begin{equation}
\left\|
\mathbf G_{\mathcal M}^{\mathrm{STE}}(\mathbf R(s))
-
\mathbf G_{\mathcal M}^{\mathrm{STE}}(\mathbf R(-s))
\right\|_F
\leq
L\|\mathbf R(s)-\mathbf R(-s)\|_F
\end{equation}
for all sufficiently small $s>0$. Taking $s\downarrow 0$, the right-hand
side converges to $0$, while the left-hand side converges to
$\Delta\|\mathbf v\|_F>0$, which is a contradiction. Therefore,
$\mathbf G_{\mathcal M}^{\mathrm{STE}}$ is not locally Lipschitz continuous at
$\mathbf R_0$, and hence cannot be globally Lipschitz continuous on
$\mathcal M$.
\end{proof}
\subsection{Proof of Proposition 2}\label{app:a.2}
\begin{proof}
Since $\widehat{G}_t\in T_{R_t}O(n)$, there exists a skew-symmetric matrix
$A_t^\top=-A_t$ such that
\begin{equation}
\widehat{G}_t=A_tR_t .
\end{equation}
Therefore,
\begin{equation}
\Omega_t
=
-\widehat{G}_tR_t^\top
=
-A_t ,
\end{equation}
which is also skew-symmetric. Since $R_t$ is orthogonal, right multiplication by
$R_t^\top$ preserves the Frobenius norm, and hence
\begin{equation}
\|\Omega_t\|_F
=
\|\widehat{G}_tR_t^\top\|_F
=
\|\widehat{G}_t\|_F .
\label{eq:norm-preserve}
\end{equation}

By the decomposition \eqref{eq:ste-decomposition}, we have
\begin{equation}
\|\widehat{G}_t\|_F^2
=
\|H_t\|_F^2
+
2\langle H_t,\varepsilon_t\rangle
+
\|\varepsilon_t\|_F^2 .
\end{equation}
Taking conditional expectation with respect to $\mathcal{F}_t$ and using
\eqref{eq:residual-noncancellation}, we obtain
\begin{align}
\mathbb{E}\!\left[
\|\widehat{G}_t\|_F^2 \mid \mathcal{F}_t
\right]
&=
\begin{aligned}[t]
&\|H_t\|_F^2
+
2\mathbb{E}\!\left[
\langle H_t,\varepsilon_t\rangle \mid \mathcal{F}_t
\right] \\
&+
\mathbb{E}\!\left[
\|\varepsilon_t\|_F^2 \mid \mathcal{F}_t
\right]
\end{aligned} \\
&\ge
\begin{aligned}[t]
&\|H_t\|_F^2
+
(1-\kappa)
\mathbb{E}\!\left[
\|\varepsilon_t\|_F^2 \mid \mathcal{F}_t
\right]
\end{aligned} \\
&\ge
(1-\kappa)\sigma^2 .
\end{align}
where the last inequality follows from \eqref{eq:residual-floor}. This proves
\eqref{eq:estimator-floor}.

Next, from the Cayley update \eqref{eq:cayley-update-revised},
\begin{align}
R_{t+1}-R_t
&=
\left[
\left(I-\frac{\alpha_t}{2}\Omega_t\right)^{-1}
\left(I+\frac{\alpha_t}{2}\Omega_t\right)
-
I
\right]R_t  \\
&=
\alpha_t
\left(I-\frac{\alpha_t}{2}\Omega_t\right)^{-1}
\Omega_t R_t .
\end{align}
Again, since right multiplication by $R_t$ preserves the Frobenius norm,
\begin{equation}
\|R_{t+1}-R_t\|_F^2
=
\alpha_t^2
\left\|
\left(I-\frac{\alpha_t}{2}\Omega_t\right)^{-1}
\Omega_t
\right\|_F^2 .
\label{eq:disp-exact}
\end{equation}

Because $\Omega_t$ is skew-symmetric, its nonzero eigenvalues are purely
imaginary and occur in conjugate pairs. Equivalently, if
$s_i(\Omega_t)$ denotes the singular values of $\Omega_t$, then the singular
values of
\begin{equation}
\left(I-\frac{\alpha_t}{2}\Omega_t\right)^{-1}\Omega_t
\end{equation}
are
\begin{equation}
\frac{s_i(\Omega_t)}
{\sqrt{1+(\alpha_t s_i(\Omega_t)/2)^2}} .
\end{equation}
Using $\|\Omega_t\|_2\le G$ and $\alpha_t\le\overline{\alpha}$, we get
\begin{align}
\left\|
\left(I-\frac{\alpha_t}{2}\Omega_t\right)^{-1}
\Omega_t
\right\|_F^2
&=
\sum_i
\frac{s_i(\Omega_t)^2}
{1+(\alpha_t s_i(\Omega_t)/2)^2}  \\
&\ge
\frac{1}{1+(\overline{\alpha}G/2)^2}
\sum_i s_i(\Omega_t)^2  \\
&=
\frac{1}{1+(\overline{\alpha}G/2)^2}
\|\Omega_t\|_F^2 .
\end{align}
Combining this inequality with \eqref{eq:disp-exact} and
\eqref{eq:norm-preserve}, we obtain
\begin{equation}
\|R_{t+1}-R_t\|_F^2
\ge
\frac{\alpha_t^2}{1+(\overline{\alpha}G/2)^2}
\|\widehat{G}_t\|_F^2 .
\end{equation}
Taking conditional expectation and applying \eqref{eq:estimator-floor} yields
\eqref{eq:cayley-displacement-lower}.

If $\alpha_t\ge\underline{\alpha}>0$, then taking total expectation and the
limit inferior gives \eqref{eq:nonvanishing-displacement}. If instead
$\alpha_t\to0$, dividing \eqref{eq:cayley-displacement-lower} by
$\alpha_t^2$ and then taking total expectation and the limit inferior yields
\eqref{eq:normalized-displacement}. This completes the proof.
\end{proof}

\subsection{Proof of Lemma 1}\label{app:a.3}
\begin{proof}
If $r=0$, then $V=0$ and the claim is trivial. Assume $r>0$.

For any $G\in O(2)$, orthogonality gives
\begin{equation}
\|VG\|_2=\|V\|_2=r.
\end{equation}
For any $x\in\mathbb{R}^2$,
\begin{equation}
\|x\|_\infty \ge \frac{\|x\|_2}{\sqrt{2}}.
\end{equation}
Therefore,
\begin{equation}
\|VG\|_\infty \ge \frac{\|VG\|_2}{\sqrt{2}}
=\frac{r}{\sqrt{2}},
\end{equation}
which proves
\begin{equation}
\min_{G\in O(2)}\|VG\|_\infty \ge \frac{r}{\sqrt{2}}.
\end{equation}

It remains to show that this lower bound is attainable. Let
\begin{equation}
\phi=\operatorname{atan2}(b,a),
\end{equation}
so that
\begin{equation}
a=r\cos\phi,\qquad b=r\sin\phi.
\end{equation}
For the Givens rotation $G(\theta)$, we have
\begin{equation}
VG(\theta)
=
\left(
a\cos\theta+b\sin\theta,\,
b\cos\theta-a\sin\theta
\right).
\end{equation}
Substituting $a=r\cos\phi$ and $b=r\sin\phi$ gives
\begin{equation}
VG(\theta)
=
\left(
r\cos(\phi-\theta),\,
r\sin(\phi-\theta)
\right).
\end{equation}
Choosing
\begin{equation}
\theta^\star=\phi-\frac{\pi}{4}
=
\operatorname{atan2}(b,a)-\frac{\pi}{4},
\end{equation}
we obtain
\begin{equation}
VG(\theta^\star)
=
\left(
r\cos\frac{\pi}{4},\,
r\sin\frac{\pi}{4}
\right)
=
\left(
\frac{r}{\sqrt{2}},
\frac{r}{\sqrt{2}}
\right).
\end{equation}
Hence
\begin{equation}
\|VG(\theta^\star)\|_\infty=\frac{r}{\sqrt{2}}.
\end{equation}
Combining this attainability with the lower bound proves
\begin{equation}
\min_{G\in O(2)}\|VG\|_\infty=\frac{r}{\sqrt{2}}.
\end{equation}
\end{proof}

\section{Additional Experiments}
\begin{figure}
	\centering
	\includegraphics[width=0.5\textwidth]{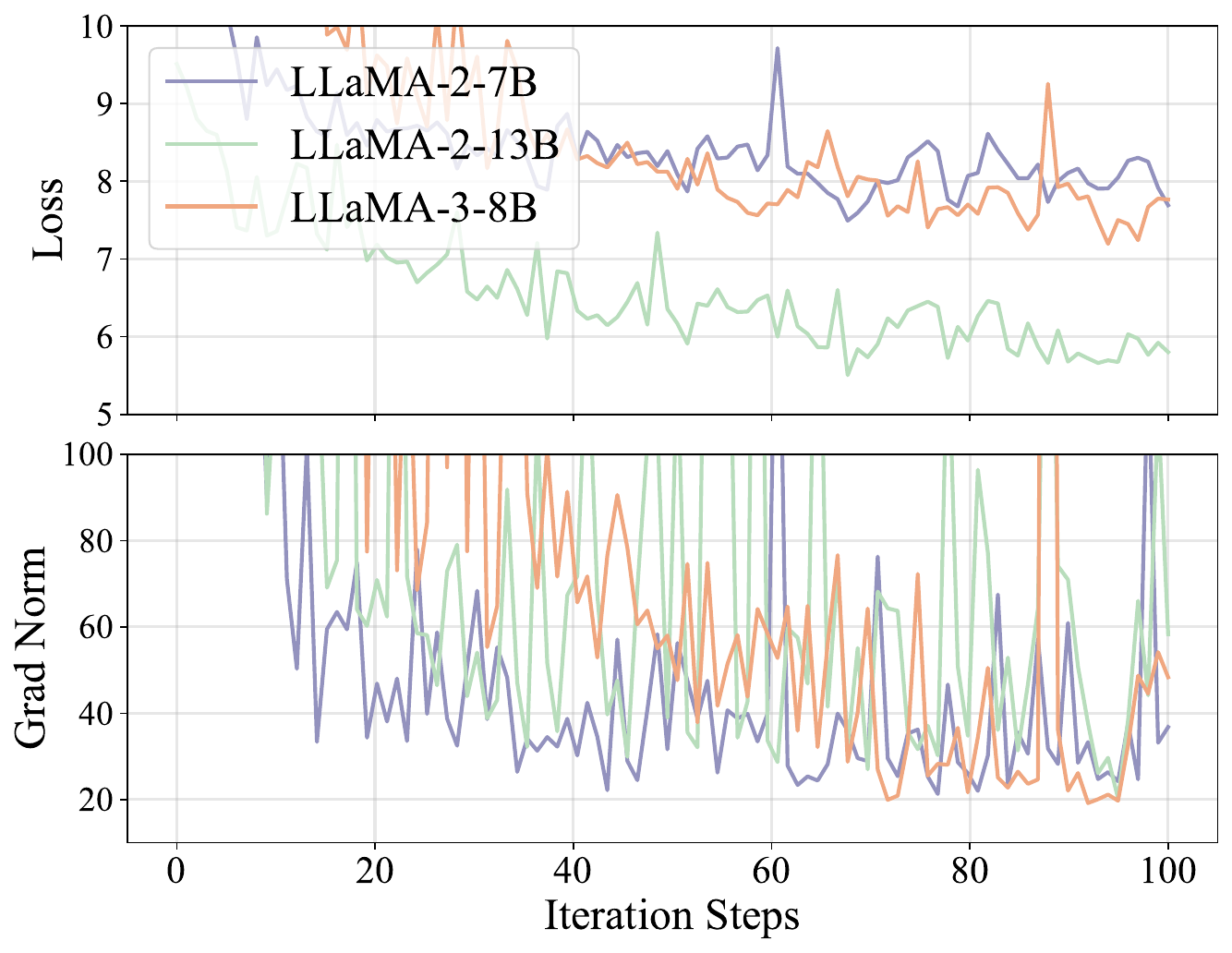} 
	\caption{The figure reports the variations in loss and gradient norms of SpinQuant across different models, with iterations fixed at 100 steps—matching SpinQuant's prescribed iteration count configuration.}
	\label{app:spinquant_loss}
\end{figure}

\textbf{Supplementary Analysis of SpinQuant Optimization.}
Fig.~\ref{app:spinquant_loss} provides additional evidence for the unstable optimization
behavior discussed in Fig.~\ref{spinquant_loss}. We report the loss and gradient norm
curves of SpinQuant on LLaMA-2-7B, LLaMA-2-13B, and LLaMA-3-8B under the
prescribed 100 optimization steps. Across different model scales and
architectures, the loss curves fail to converge smoothly, while the gradient
norms exhibit persistent spikes throughout the optimization process. This
suggests that the oscillatory behavior is not an isolated case on a single
model, but a common phenomenon when STE-based rotation learning is coupled with
low-bit quantization. These observations further support our motivation for
replacing iterative Cayley-SGD rotation learning with the closed-form rotation
construction used in SingleQuant.

\textbf{Detailed Zero-shot QA Results.}
Tab.~\ref{app:tab2} provides the detailed breakdown of the six zero-shot QA
benchmarks summarized in Tab.~\ref{tab2}. Across ARC-C, ARC-E, HellaSwag,
LAMBADA, PIQA, and WinoGrande, SingleQuant consistently preserves strong task
performance under W4A4 quantization. In particular, SingleQuant achieves the
best average accuracy among quantized methods on LLaMA-2-7B, LLaMA-2-13B,
LLaMA-2-70B, and LLaMA-3-70B, while remaining competitive on LLaMA-3-8B. The
per-task results show that the improvements are not driven by a single dataset;
rather, SingleQuant maintains balanced accuracy across commonsense reasoning,
reading comprehension, and language modeling-oriented QA tasks. These detailed
results further support the conclusion in Tab.~\ref{tab2} that the proposed
ART and URT components effectively reduce the accuracy degradation caused by
low-bit weight-activation quantization.

\begin{table*}[h]
	\centering
	\caption{Zero-shot QA task results of 4-bit weight \& activation quantized LLaMA models.}
	
	\begin{adjustbox}{width=\linewidth,center}
		\begin{tabular}{@{}llcccccccc@{}}
			\toprule
			Model & Method & W Quant. & ARC-C & ARC-E & HellaSwag & LAMBADA & PIQA & Winogrande & Avg.\textuparrow \\
			\midrule
			\multirow{7}{*}{2-7B} 
			& FP16 & - & 46.42 & 74.33 & 75.98 & 73.92 & 79.05 & 69.53 & 69.87 \\
			& QuaRot & RTN & 36.60 & 61.41 & 65.07 & 48.06 & 72.20 & 63.06 & 57.73 \\
			& SpinQuant & RTN & 39.42 & 65.32 & 71.45 & 66.16 & 75.30 & 63.46 & 63.52 \\
			&DuQuant & RTN & 38.57 & 51.26 & 69.39 & 69.63 & 75.95 & 63.22 & 61.34 \\
			& QuaRot & GPTQ & 42.32 & 68.35 & 72.53 & 65.40 & 76.33 & 65.11 & 65.01 \\
			& SpinQuant & GPTQ & 41.72 & 69.28 & 72.90 & 71.28 & 76.17 & 66.06 & 66.23 \\
			&OSTQuant&GPTQ&41.64&68.94&\textbf{73.17}&\textbf{71.61}&\textbf{77.09}&65.90&66.39\\
			& \pinkcell{\textbf{SingleQuant}} & \pinkcell{RTN} & \pinkcell{\textbf{42.92}} & \pinkcell{\textbf{70.08}} & \pinkcell{72.86} & \pinkcell{71.10} & \pinkcell{76.83} & \pinkcell{\textbf{69.30}} & \pinkcell{\textbf{67.18}} \\
			\midrule
			
			\multirow{7}{*}{2-13B}
			& FP16 & - & 49.15 & 77.44 & 79.39 & 76.73 & 80.47 & 72.14 & 72.55 \\
			& QuaRot & RTN & 42.83 & 69.95 & 73.54 & 65.62 & 77.69 & 67.88 & 66.25 \\
			& SpinQuant & RTN & 43.69 & 72.43 & 75.52 & 72.42 & 78.40 & 68.90 & 68.56 \\
			& DuQuant & RTN & 43.26 & 56.02 & 73.87 & 73.92 & 77.31 & 65.51 & 64.98 \\
			& QuaRot & GPTQ & 45.48 & 73.27 & 76.03 & 69.01 & 79.05 & 70.64 & 68.91 \\
			& SpinQuant & GPTQ & \textbf{49.15} & \textbf{77.19} & 76.86 & 73.86 & 78.67 & 69.85 & 70.93 \\
			&OSTQuant&GPTQ&47.10&75.20&\textbf{77.46}&\textbf{75.14}&78.67&68.03&70.27\\
			& \pinkcell{\textbf{SingleQuant}} & \pinkcell{RTN} & \pinkcell{49.02} & \pinkcell{76.96} & \pinkcell{77.34 }& \pinkcell{74.93} &\pinkcell{\textbf{79.94}} & \pinkcell{\textbf{70.79}} & \pinkcell{\textbf{71.50}} \\
			\midrule
			
			\multirow{7}{*}{2-70B}
			& FP16 & - & 57.17 & 81.02 & 83.81 & 79.60 & 82.70 & 77.98 & 77.05 \\
			& QuaRot & RTN & 52.22 & 76.60 & 79.96 & 74.61 & 81.12 & 76.32 & 73.47 \\
			& SpinQuant & RTN & 55.03 & 79.17 & 81.76 & 78.87 & 81.45 & 74.27 & 75.09 \\
			& DuQuant & RTN & 46.67 & 58.88 & 79.62 & 76.44 & 79.76 & 74.98 & 69.39 \\
			& QuaRot & GPTQ & 55.46 & 79.76 & 81.58 & \textbf{79.35} & 81.83 & 76.09 & 75.68 \\
			& SpinQuant & GPTQ & 55.38 & 79.04 & 82.57 & 78.75 & 82.37 & \textbf{78.22} & 76.06 \\
			& \pinkcell{\textbf{SingleQuant}} & \pinkcell{RTN} & \pinkcell{\textbf{56.57}} & \pinkcell{\textbf{79.88}} & \pinkcell{\textbf{82.69}} & \pinkcell{78.53} & \pinkcell{\textbf{82.68}} & \pinkcell{76.48} & \pinkcell{\textbf{76.14}} \\
			\midrule
			
			\multirow{7}{*}{3-8B}
			& FP16 & - & 53.50 & 77.57 & 79.12 & 75.51 & 80.74 & 72.93 & 73.23 \\
			& QuaRot & RTN & 38.65 & 66.54 & 68.82 & 57.20 & 71.82 & 65.04 & 61.34 \\
			& SpinQuant & RTN & 45.73 & 71.38 & 74.07 & 67.67 & 76.66 & 66.38 & 66.98 \\
			& DuQuant & RTN & 41.81 & 68.48 & 73.07 & 69.32 & 75.68 & 66.22 & 65.76\\
			& QuaRot & GPTQ & 45.73 & 70.83 & 72.97 & 62.70 & 75.35 & 67.17 & 65.79 \\
			& SpinQuant & GPTQ &47.27 & 74.20 & 74.55 & 70.29 & 77.37& 68.51 & 68.70 \\
			&OSTQuant&GPTQ&\textbf{ 48.21}&72.69&75.69&70.52&\textbf{ 77.86}&\textbf{ 69.53}&\textbf{69.08}\\
			& \pinkcell{\textbf{SingleQuant}} & \pinkcell{RTN} & \pinkcell{46.19} &\pinkcell{\textbf{75.71}} & \pinkcell{\textbf{76.03}} & \pinkcell{\textbf{71.02}} & \pinkcell{76.92} & \pinkcell{67.87} & \pinkcell{68.96} \\
			\midrule
			
			\multirow{7}{*}{3-70B}
			& FP16 & - & 64.25 & 85.94 & 84.93 & 79.37 & 84.44 & 80.74 & 79.95 \\
			& QuaRot & RTN & 22.18 & 34.30 & 32.15 & 13.35 & 57.67 & 52.49 & 35.36 \\
			& SpinQuant & RTN & 44.03 & 69.07 & 74.57 & 63.34 & 76.99 & 65.98 & 65.66 \\
			& DuQuant & RTN & \textbf{58.87} & 82.58 & 74.94 & 70.29 & \textbf{82.70} & 68.43 & 72.97 \\
			& QuaRot & GPTQ & 44.19 & 69.33 & 74.54 & 63.47 & 77.30 & 66.41 & 65.37 \\
			& SpinQuant & GPTQ & 50.51 & 75.88 & 76.49 & 73.20 & 79.00 & 72.93 & 71.33 \\
			& \pinkcell{\textbf{SingleQuant}} & \pinkcell{RTN} & \pinkcell{58.19} & \pinkcell{\textbf{82.62}} & \pinkcell{\textbf{82.53}} & \pinkcell{\textbf{76.67}} & \pinkcell{81.77} & \pinkcell{\textbf{76.01}} & \pinkcell{\textbf{76.30}} \\
			\bottomrule
		\end{tabular}
	\end{adjustbox}
	\label{app:tab2}
\end{table*}

\textbf{Quantization Time of SingleQuant.}
Tab.~\ref{runtime} reports the standalone quantization time of SingleQuant on a
single NVIDIA A800 80GB GPU, averaged over 30 independent runs. SingleQuant
finishes quantizing medium-scale models within one minute, requiring only 24s
for LLaMA-2-7B, 37s for LLaMA-2-13B, and 27s for LLaMA-3-8B. Even for 70B-scale
models, the total quantization time remains within a few minutes, i.e.,
4.5 minutes for LLaMA-2-70B and 3.3 minutes for LLaMA-3-70B. These results
show that the closed-form ART/URT construction and Kronecker decomposition
avoid the heavy iterative optimization cost required by learnable rotation
methods. Therefore, SingleQuant provides scalable quantization efficiency
across different model sizes while preserving the practicality of W4A4
deployment.

\begin{table}[htbp]
	\centering
	\caption{The quantization time results for SingleQuant running on a single NVIDIA A800 80G GPU were measured, with each data point independently executed 10 times.}
	\renewcommand\tabcolsep{4.0pt}
	\begin{tabular}{c|ccccc}
		\toprule
		SingleQuant& 2-7B&2-13B&2-70B&3-8B&3-70B\\
		\midrule
		Quantization Time&24s&37s&4.5min&27s&3.3min\\
		\bottomrule
	\end{tabular}
	\label{app:runtime}
\end{table}

\textbf{Weight-only Quantization.}
Table B.3 evaluates SingleQuant under weight-only quantization on LLaMA-3-8B. Although designed for W4A4 quantization, SingleQuant also remains effective when only weights are quantized. Under W4A16, it achieves 6.40 perplexity on WikiText-2 and 9.57 on C4, closely approaching FP16 and outperforming all compared baselines. Under the more challenging W3A16 setting, SingleQuant maintains competitive perplexity of 7.87 on WikiText-2 and 12.31 on C4, while RTN and several optimization-based methods suffer substantial degradation. These results show that the proposed closed-form rotation provides a general distribution-smoothing effect beyond weight-activation quantization.

\begin{table}[htbp]
	\centering
	\caption{WikiText-2 and C4 perplexity of weight-only quantizationon on LLaMA-3-8B model.}
	\begin{tabular}{c|cc|cc}
		\toprule
		\multirow{2}{*}{LLaMA-3-8B} & \multicolumn{2}{c|}{WikiText-2 } & \multicolumn{2}{c}{C4} \\
		\cmidrule{2-5}
		& W4A16 & W3A16 & W4A16 & W3A16 \\
		FP16 & \multicolumn{2}{c|}{6.14} & \multicolumn{2}{c}{9.45} \\
		\midrule

		RTN & 8.70 & 2.2E3 & 14.00 & 5.6E3 \\

		GPTQ & 7.00 & 13.00 & 11.80 & 45.90 \\
		GPTQ-g128 & 6.50 & 8.20 & 10.40 & 13.70 \\
		AWQ & 7.10 & 12.80 & 10.10 & 16.80 \\
		QuIP & 6.50 & 7.50 & 11.10 & 11.30 \\
		\textbf{SingleQuant} & 6.40 & 7.87 & 9.57 & 12.31 \\
		\bottomrule
	\end{tabular}
	\label{wquant}
\end{table}

\printcredits

\bibliographystyle{cas-model2-names}

\bibliography{cas-refs}

@article{lin2024duquant,
  title={Duquant: Distributing outliers via dual transformation makes stronger quantized llms},
  author={Lin, Haokun and Xu, Haobo and Wu, Yichen and Cui, Jingzhi and Zhang, Yingtao and Mou, Linzhan and Song, Linqi and Sun, Zhenan and Wei, Ying},
  journal={Advances in Neural Information Processing Systems},
  volume={37},
  pages={87766--87800},
  year={2024}
}

@inproceedings{xiao2023smoothquant,
  title={Smoothquant: Accurate and efficient post-training quantization for large language models},
  author={Xiao, Guangxuan and Lin, Ji and Seznec, Mickael and Wu, Hao and Demouth, Julien and Han, Song},
  booktitle={International Conference on Machine Learning},
  pages={38087--38099},
  year={2023},
  organization={PMLR}
}

@inproceedings{shaoomniquant,
  title={OmniQuant: Omnidirectionally Calibrated Quantization for Large Language Models},
  author={Shao, Wenqi and Chen, Mengzhao and Zhang, Zhaoyang and Xu, Peng and Zhao, Lirui and Li, Zhiqian and Zhang, Kaipeng and Gao, Peng and Qiao, Yu and Luo, Ping},
  booktitle={The Twelfth International Conference on Learning Representations},
  year={2024}
}

@article{bengio2013estimating,
  title={Estimating or propagating gradients through stochastic neurons for conditional computation},
  author={Bengio, Yoshua and L{\'e}onard, Nicholas and Courville, Aaron},
  journal={arXiv preprint arXiv:1308.3432},
  year={2013}
}

@inproceedings{liefficient,
  title={Efficient Riemannian Optimization on the Stiefel Manifold via the Cayley Transform},
  author={Li, Jun and Li, Fuxin and Todorovic, Sinisa},
  booktitle={International Conference on Learning Representations},
year={2020}
}

@book{press2007numerical,
  title={Numerical recipes 3rd edition: The art of scientific computing},
  author={Press, William H},
  year={2007},
  publisher={Cambridge university press}
}

@inproceedings{liu2024spinquant,
  title={SpinQuant: LLM Quantization with Learned Rotations},
  author={Liu, Zechun and Zhao, Changsheng and Fedorov, Igor and Soran, Bilge and Choudhary, Dhruv and Krishnamoorthi, Raghuraman and Chandra, Vikas and Tian, Yuandong and Blankevoort, Tijmen},
  booktitle={The Thirteenth International Conference on Learning Representations},
  year={2024}
}

@inproceedings{ma2024parameter,
  title={Parameter efficient quasi-orthogonal fine-tuning via givens rotation},
  author={Ma, Xinyu and Chu, Xu and Yang, Zhibang and Lin, Yang and Gao, Xin and Zhao, Junfeng},
  booktitle={Proceedings of the 41st International Conference on Machine Learning},
  pages={33686--33729},
  year={2024}
}

@article{touvron2023llama,
  title={Llama 2: Open foundation and fine-tuned chat models},
  author={Touvron, Hugo and Martin, Louis and Stone, Kevin and Albert, Peter and Almahairi, Amjad and Babaei, Yasmine and Bashlykov, Nikolay and Batra, Soumya and Bhargava, Prajjwal and Bhosale, Shruti and others},
  journal={arXiv preprint arXiv:2307.09288},
  year={2023}
}

@article{grattafiori2024llama,
  title={The llama 3 herd of models},
  author={Grattafiori, Aaron and Dubey, Abhimanyu and Jauhri, Abhinav and Pandey, Abhinav and Kadian, Abhishek and Al-Dahle, Ahmad and Letman, Aiesha and Mathur, Akhil and Schelten, Alan and Vaughan, Alex and others},
  journal={arXiv preprint arXiv:2407.21783},
  year={2024}
}

@inproceedings{merity2017pointer,
  title={Pointer Sentinel Mixture Models},
  author={Merity, Stephen and Xiong, Caiming and Bradbury, James and Socher, Richard},
  booktitle={International Conference on Learning Representations},
  year={2017}
}

@article{raffel2020exploring,
  title={Exploring the limits of transfer learning with a unified text-to-text transformer},
  author={Raffel, Colin and Shazeer, Noam and Roberts, Adam and Lee, Katherine and Narang, Sharan and Matena, Michael and Zhou, Yanqi and Li, Wei and Liu, Peter J},
  journal={Journal of machine learning research},
  volume={21},
  number={140},
  pages={1--67},
  year={2020}
}

@article{clark2018think,
  title={Think you have solved question answering? try arc, the ai2 reasoning challenge},
  author={Clark, Peter and Cowhey, Isaac and Etzioni, Oren and Khot, Tushar and Sabharwal, Ashish and Schoenick, Carissa and Tafjord, Oyvind},
  journal={arXiv preprint arXiv:1803.05457},
  year={2018}
}

@inproceedings{zellers2019hellaswag,
  title={HellaSwag: Can a Machine Really Finish Your Sentence?},
  author={Zellers, Rowan and Holtzman, Ari and Bisk, Yonatan and Farhadi, Ali and Choi, Yejin},
  booktitle={Proceedings of the 57th Annual Meeting of the Association for Computational Linguistics},
  pages={4791--4800},
  year={2019}
}

@inproceedings{paperno2016lambada,
  title={The LAMBADA dataset: Word prediction requiring a broad discourse context},
  author={Paperno, Denis and Kruszewski, Germ{\'a}n and Lazaridou, Angeliki and Pham, Ngoc-Quan and Bernardi, Raffaella and Pezzelle, Sandro and Baroni, Marco and Boleda, Gemma and Fern{\'a}ndez, Raquel},
  booktitle={Proceedings of the 54th Annual Meeting of the Association for Computational Linguistics (Volume 1: Long Papers)},
  pages={1525--1534},
  year={2016}
}

@inproceedings{bisk2020piqa,
  title={Piqa: Reasoning about physical commonsense in natural language},
  author={Bisk, Yonatan and Zellers, Rowan and Gao, Jianfeng and Choi, Yejin and others},
  booktitle={Proceedings of the AAAI conference on artificial intelligence},
  volume={34},
  number={05},
  pages={7432--7439},
  year={2020}
}

@article{sakaguchi2021winogrande,
  title={Winogrande: An adversarial winograd schema challenge at scale},
  author={Sakaguchi, Keisuke and Bras, Ronan Le and Bhagavatula, Chandra and Choi, Yejin},
  journal={Communications of the ACM},
  volume={64},
  number={9},
  pages={99--106},
  year={2021},
  publisher={ACM New York, NY, USA}
}

@inproceedings{maaffinequant,
  title={AffineQuant: Affine Transformation Quantization for Large Language Models},
  author={Ma, Yuexiao and Li, Huixia and Zheng, Xiawu and Ling, Feng and Xiao, Xuefeng and Wang, Rui and Wen, Shilei and Chao, Fei and Ji, Rongrong},
  booktitle={The Twelfth International Conference on Learning Representations},
  year={2024}
}

@article{ashkboos2023towards,
  title={Towards end-toend 4-bit inference on generative large language models},
  author={Ashkboos, S and Markov, I and Frantar, E and Zhong, T and Wang, X and Ren, J and Hoefler, T and Alistarh, D},
  journal={arXiv preprint arXiv:2310.09259},
  year={2023}
}

@article{ashkboos2024quarot,
  title={Quarot: Outlier-free 4-bit inference in rotated llms},
  author={Ashkboos, Saleh and Mohtashami, Amirkeivan and Croci, Maximilian L and Li, Bo and Cameron, Pashmina and Jaggi, Martin and Alistarh, Dan and Hoefler, Torsten and Hensman, James},
  journal={Advances in Neural Information Processing Systems},
  volume={37},
  pages={100213--100240},
  year={2024}
}

@article{zhao2024atom,
  title={Atom: Low-bit quantization for efficient and accurate llm serving},
  author={Zhao, Yilong and Lin, Chien-Yu and Zhu, Kan and Ye, Zihao and Chen, Lequn and Zheng, Size and Ceze, Luis and Krishnamurthy, Arvind and Chen, Tianqi and Kasikci, Baris},
  journal={Proceedings of Machine Learning and Systems},
  volume={6},
  pages={196--209},
  year={2024}
}

@misc{eval-harness,
  author       = {Gao, Leo and Tow, Jonathan and Abbasi, Baber and Biderman, Stella and Black, Sid and DiPofi, Anthony and Foster, Charles and Golding, Laurence and Hsu, Jeffrey and Le Noac'h, Alain and Li, Haonan and McDonell, Kyle and Muennighoff, Niklas and Ociepa, Chris and Phang, Jason and Reynolds, Laria and Schoelkopf, Hailey and Skowron, Aviya and Sutawika, Lintang and Tang, Eric and Thite, Anish and Wang, Ben and Wang, Kevin and Zou, Andy},
  title        = {The Language Model Evaluation Harness},
  month        = 07,
  year         = 2024,
  publisher    = {Zenodo},
  version      = {v0.4.3},
  doi          = {10.5281/zenodo.12608602},
  url          = {https://zenodo.org/records/12608602}
}

@article{wolf2019huggingface,
  title={Huggingface's transformers: State-of-the-art natural language processing},
  author={Wolf, Thomas and Debut, Lysandre and Sanh, Victor and Chaumond, Julien and Delangue, Clement and Moi, Anthony and Cistac, Pierric and Rault, Tim and Louf, R{\'e}mi and Funtowicz, Morgan and others},
  journal={arXiv preprint arXiv:1910.03771},
  year={2019}
}

@article{paszke2019pytorch,
  title={Pytorch: An imperative style, high-performance deep learning library},
  author={Paszke, A},
  journal={arXiv preprint arXiv:1912.01703},
  year={2019}
}

@article{huang2024good,
  title={How good are low-bit quantized llama3 models? an empirical study},
  author={Huang, Wei and Ma, Xudong and Qin, Haotong and Zheng, Xingyu and Lv, Chengtao and Chen, Hong and Luo, Jie and Qi, Xiaojuan and Liu, Xianglong and Magno, Michele},
  journal={arXiv e-prints},
  pages={arXiv--2404},
  year={2024}
}

@article{huang2025large,
  title={Are large language models qualified reviewers in originality evaluation?},
  author={Huang, Shengzhi and Huang, Yong and Liu, Yinpeng and Luo, Zhuoran and Lu, Wei},
  journal={Information Processing \& Management},
  volume={62},
  number={3},
  pages={103973},
  year={2025},
  publisher={Elsevier}
}

@article{aggarwal2026large,
  title={Large language models for scholarly ontology generation: An extensive analysis in the engineering field},
  author={Aggarwal, Tanay and Salatino, Angelo and Osborne, Francesco and Motta, Enrico},
  journal={Information Processing \& Management},
  volume={63},
  number={1},
  pages={104262},
  year={2026},
  publisher={Elsevier}
}

@inproceedings{jinmassive,
  title={Massive Values in Self-Attention Modules are the Key to Contextual Knowledge Understanding},
  author={Jin, Mingyu and Mei, Kai and Xu, Wujiang and Sun, Mingjie and Tang, Ruixiang and Du, Mengnan and Liu, Zirui and Zhang, Yongfeng},
  booktitle={Forty-second International Conference on Machine Learning},
  year={2025}
}

@inproceedings{sunflatquant,
  title={FlatQuant: Flatness Matters for LLM Quantization},
  author={Sun, Yuxuan and Liu, Ruikang and Bai, Haoli and Bao, Han and Zhao, Kang and Li, Yuening and Yu, Xianzhi and Hou, Lu and Yuan, Chun and Jiang, Xin and others},
  booktitle={Forty-second International Conference on Machine Learning},
  year={2025}
}

@inproceedings{huostquant,
  title={OSTQuant: Refining Large Language Model Quantization with Orthogonal and Scaling Transformations for Better Distribution Fitting},
  author={Hu, Xing and Cheng, Yuan and Yang, Dawei and Chen, Zhixuan and Xu, Zukang and Yuan, Zhihang and Zhou, Sifan and others},
  booktitle={The Thirteenth International Conference on Learning Representations},
  year={2025}
}

@inproceedings{frantaroptq,
  title={OPTQ: Accurate Quantization for Generative Pre-trained Transformers},
  author={Frantar, Elias and Ashkboos, Saleh and Hoefler, Torsten and Alistarh, Dan},
  booktitle={The Eleventh International Conference on Learning Representations},
  year={2023}
}

@article{chiang2023vicuna,
  title={Vicuna: An open-source chatbot impressing gpt-4 with 90\%* chatgpt quality},
  author={Chiang, Wei-Lin and Li, Zhuohan and Lin, Ziqing and Sheng, Ying and Wu, Zhanghao and Zhang, Hao and Zheng, Lianmin and Zhuang, Siyuan and Zhuang, Yonghao and Gonzalez, Joseph E and others},
  journal={See https://vicuna. lmsys. org (accessed 14 April 2023)},
  volume={2},
  number={3},
  pages={6},
  year={2023}
}

@article{yin2019understanding,
  title={Understanding straight-through estimator in training activation quantized neural nets},
  author={Yin, Penghang and Lyu, Jiancheng and Zhang, Shuai and Osher, Stanley and Qi, Yingyong and Xin, Jack},
  journal={arXiv preprint arXiv:1903.05662},
  year={2019}
}

@article{wei2022outlier,
  title={Outlier suppression: Pushing the limit of low-bit transformer language models},
  author={Wei, Xiuying and Zhang, Yunchen and Zhang, Xiangguo and Gong, Ruihao and Zhang, Shanghang and Zhang, Qi and Yu, Fengwei and Liu, Xianglong},
  journal={Advances in Neural Information Processing Systems},
  volume={35},
  pages={17402--17414},
  year={2022}
}

@inproceedings{liu2024intactkv,
  title={IntactKV: Improving Large Language Model Quantization by Keeping Pivot Tokens Intact},
  author={Liu, Ruikang and Bai, Haoli and Lin, Haokun and Li, Yuening and Gao, Han and Xu, Zhengzhuo and Hou, Lu and Yao, Jun and Yuan, Chun},
  booktitle={Findings of the Association for Computational Linguistics ACL 2024},
  pages={7716--7741},
  year={2024}
}

@inproceedings{sun2024massive,
  title={Massive Activations in Large Language Models},
  author={Sun, Mingjie and Chen, Xinlei and Kolter, J Zico and Liu, Zhuang},
  booktitle={First Conference on Language Modeling},
  year={2024}
}

@inproceedings{bai2021binarybert,
  title={BinaryBERT: Pushing the Limit of BERT Quantization},
  author={Bai, Haoli and Zhang, Wei and Hou, Lu and Shang, Lifeng and Jin, Jin and Jiang, Xin and Liu, Qun and Lyu, Michael and King, Irwin},
  booktitle={Proceedings of the 59th Annual Meeting of the Association for Computational Linguistics and the 11th International Joint Conference on Natural Language Processing (Volume 1: Long Papers)},
  pages={4334--4348},
  year={2021}
}

@inproceedings{shen2020q,
  title={Q-bert: Hessian based ultra low precision quantization of bert},
  author={Shen, Sheng and Dong, Zhen and Ye, Jiayu and Ma, Linjian and Yao, Zhewei and Gholami, Amir and Mahoney, Michael W and Keutzer, Kurt},
  booktitle={Proceedings of the AAAI conference on artificial intelligence},
  volume={34},
  number={05},
  pages={8815--8821},
  year={2020}
}

@inproceedings{hendrycksmeasuring,
  title={Measuring Massive Multitask Language Understanding},
  author={Hendrycks, Dan and Burns, Collin and Basart, Steven and Zou, Andy and Mazeika, Mantas and Song, Dawn and Steinhardt, Jacob},
  booktitle={International Conference on Learning Representations},
  year={2021}
}

@article{wang2025mergequant,
  title={MergeQuant: Accurate 4-bit Static Quantization of Large Language Models by Channel-wise Calibration},
  author={Wang, Jinguang and Wang, Jingyu and Sun, Haifeng and Yang, Tingting and Zhuang, Zirui and Ning, Wanyi and Yin, Yuexi and Qi, Qi and Liao, Jianxin},
  journal={arXiv preprint arXiv:2503.07654},
  year={2025}
}

@inproceedings{barberoround,
  title={Round and Round We Go! What makes Rotary Positional Encodings useful?},
  author={Barbero, Federico and Vitvitskyi, Alex and Perivolaropoulos, Christos and Pascanu, Razvan and Veli{\v{c}}kovi{\'c}, Petar},
  booktitle={The Thirteenth International Conference on Learning Representations},
year={2025}
}

@inproceedings{liu2025fbquant,
  title={FBQuant: feedback quantization for large language models},
  author={Liu, Yijiang and Fang, Hengyu and He, Liulu and Zhang, Rongyu and Bai, Yichuan and Du, Yuan and Du, Li},
  booktitle={Proceedings of the Thirty-Fourth International Joint Conference on Artificial Intelligence},
  pages={7589--7597},
  year={2025}
}

@inproceedings{ramachandran2025microscopiq,
  title={Microscopiq: Accelerating foundational models through outlier-aware microscaling quantization},
  author={Ramachandran, Akshat and Kundu, Souvik and Krishna, Tushar},
  booktitle={Proceedings of the 52nd Annual International Symposium on Computer Architecture},
  pages={1193--1209},
  year={2025}
}



\end{document}